\documentclass[letterpaper]{article} 
\usepackage[preprint]{aaai2027}  

\PassOptionsToPackage{numbers, compress}{natbib}

\usepackage[hyphens]{url}  
\usepackage{graphicx} 
\usepackage{natbib}  
\usepackage{caption} 
\usepackage{algorithm}
\usepackage{algorithmic}

\usepackage{newfloat}
\usepackage{listings}
\DeclareCaptionStyle{ruled}{labelfont=normalfont,labelsep=colon,strut=off} 
\floatstyle{ruled}
\newfloat{listing}{tb}{lst}{}
\floatname{listing}{Listing}

\usepackage{booktabs}

\usepackage{hyperref}
\usepackage{array}
\usepackage{amsfonts}
\usepackage{subcaption}
\usepackage{booktabs}
\usepackage{makecell}
\usepackage{graphicx}
\usepackage{makecell}
\usepackage[table]{xcolor}
\usepackage{caption}
\usepackage{amsmath} 
\usepackage{nameref}
\usepackage{amssymb}
\usepackage{multirow}

\title{Shadow Feature Refinement Network: Progressive Feature Refinement based on Knowledge Distillation for Effective Shadow Removal}
\author{
Donghyun Han\textsuperscript{\rm 1},
Byoung-dai Lee\textsuperscript{\rm 1}\corresponding
}
\affiliations{
    \textsuperscript{\rm 1} Kyonggi University, Korea \\

    mpolio2@kyonggi.ac.kr, blee@kgu.ac.kr
}

\makeatletter
\def\ps@submissionnotice{%
  \let\@oddhead\@empty
  \let\@evenhead\@empty
  \def\@oddfoot{%
    \hfil
    {\footnotesize
     Published in \copyright~Multimedia Systems 31, article 449 (2025). {\color{magenta}\url{https://doi.org/10.1007/s00530-025-02036-5}}
     }%
    \hfil
  }%
  \let\@evenfoot\@oddfoot
}
\makeatother

\begin{document}

\nocopyright
\maketitle

\thispagestyle{submissionnotice}

\begin{abstract}
\label{sec:abs}
In the field of deep learning, has seen significant advancements; however, shadow removal remains a persistent challenge owing to the variable sizes and colors of shadows influenced by lighting conditions. This study proposes a novel shadow feature refinement network (SFR-Net), which leverages supervised learning, feature refinement loss, and knowledge distillation to enhance shadow removal performance. A dedicated post-processing algorithm is further introduced to restore natural color consistency in the generated shadow-free images. We evaluated our method on two public datasets: the adjusted image shadow triplet dataset (ISTD+) and the shadow removal dataset (SRD), which demonstrate strong generalization capabilities under diverse conditions. On ISTD+, our model achieved a root mean square error (RMSE) of 3.4627 and structural similarity index measure (SSIM) of 0.9382 across the entire image. On SRD, it recorded an RMSE of 4.3781 and an SSIM of 0.9341. These comprehensive results show that our approach performs competitively across both shadow and non-shadow regions while setting a promising direction for robust and perceptually natural shadow removal. Code is available at {\color{magenta}\url{https://github.com/DongHyun99/SFRNet}.}
\end{abstract}
\section{Introduction}
\label{sec:intro}

A shadow forms when light is obstructed by an opaque object, preventing it from reaching a surface. It is a complex phenomenon determined by various environmental factors, such as object shape, lighting conditions, and reflections~\cite{ref01,ref02,ref03}. Shadows are often disruptive in image processing tasks, such as image segmentation and object detection~\cite{ref04,ref05,ref06,ref07}, prompting several studies to propose new methods for shadow removal~\cite{ref01,ref02,ref08,ref09,ref10,ref11}. Nonetheless, shadow removal is complex and challenging, for several reasons. First, shadows cause nonuniform lighting in certain regions because of light obstruction. Therefore, removing shadows by adjusting the brightness or contrast of an image is challenging~\cite{ref12}. This is because non-uniform lighting induces changes in the texture and colors of shadow regions, resulting in the loss of image information. Second, it is difficult to distinguish shadows from actual objects~\cite{ref13}, particularly in complex scenes, making accurate detection even more difficult. Moreover, shadows consist of two regions: the umbra, where the light source is completely blocked, and the penumbra, where light source is only partially blocked. The size and sharpness of the penumbral region vary depending on the intensity and distance of the light source, which makes difficult to define the shadow boundary accurately~\cite{ref01}. The challenge is compounded in cases involving overlapping shadows formed by multiple light sources and reflections. 

The key to shadow removal lies in estimating and restoring the colors of shadowed regions based on those of nonshadowed regions. Various methods have been introduced in recent years. Classical shadow removal approaches remove shadows based on prior information regarding the lighting or gradient of an image~\cite{ref14,ref15,ref16}. However, they suffer from high computational costs and design complexity~\cite{ref10}. Recent advances in deep-learning-based generative models have enabled significant progress in shadow removal. In particular, supervised learning approaches integrate paired shadow and ground truth (GT) images, thereby allowing the model to learn the differences between the two images~\cite{ref17}. These methods~\cite{ref01,ref03,ref08,ref09,ref10,ref11,ref17,ref18,ref19,ref20,ref21,ref22} have demonstrated superior performance and accuracy than conventional methods.

However, several limitations remain. First, constructing diverse and accurate shadow datasets is labor-intensive and time-consuming~\cite{ref08}. In outdoor images captured under sunlight, minute differences often exist in pixel values of the non-shadowed regions across image pairs~\cite{ref23}. Because of the characteristics of shadow removal models that restore shadowed regions through non-shadowed regions, such differences can pose a significant hindrance to learning. However, it is difficult to obtain images where there is almost no difference between paired images. Consequently, there are
limitations in reconstructing images through a simple comparison of the two images. Second, as previously mentioned, it is difficult to construct a shadow dataset. Therefore, there is limited diversity in the objects and environments that can induce biased learning~\cite{ref23}. The trained model does not flexibly generalize to new shadow images, because the shapes and types of objects that create shadows are similar, and the extent of shadows or locations where images can be captured is limited~\cite{ref24}.

Image inpainting is the task of restoring masked-out regions based on the information gathered from the surrounding regions. Similarly, because the shadow removal task also restores the shadow region to a state where there is no shadow, the two tasks are common in that they both deal with visual imperfections. However, while image inpainting
generates entirely new content by utilizing the surrounding information, shadow removal can use ample information, including objects, textures, and colors, from both the surrounding regions and the shadow region. Hence, shadow removal can be seen as a specific form of image inpainting in a narrow sense. Therefore, this study proposes a supervised learning-based shadow removal model, referred to as the Shadow Feature Refinement network (SFR-Net), which uses the image inpainting concept. We incorporate image inpainting technology and also utilize the knowledge distillation approach to remove shadows.

Our hypothesis related to this approach is as follows: In some cases, the model is trained such that the final image generated by removing shadows from the input image is similar to the GT image of the input image. Further, in other instances, the model is trained such that the features generated during the shadow removal process are similar to the features generated when the GT image passes through the image inpainting network.

In these cases, it is possible to achieve a more realistic image with shadows removed. Therefore, the SFR-Net is composed of two image generation networks that are derived
from the image inpainting network model: guiding network ($G_I$) and guided network ($G_S$). The $G_S$ receives shadow image ($I_S$) and the shadow region of the shadow image, which is shadow mask ($M_S$), as input to remove shadows. The $G_I$ is an image generation model with the same structure as the $G_S$and utilizes the GT image ($I_{gt}$) and an empty shadow mask ($M_{gt}$) as input to reconstruct the GT image.
In this process, the features generated at each layer play the role of supervising the training of the corresponding layer in the guided network. To verify the effectiveness of the technique we proposed, we conducted extensive experiments using the ISTD+~\cite{ref08} dataset. The experimental results indicate the SFR-Net yielded a competitive performance, both quantitatively and qualitatively, compared to recent techniques,  including Shadow Former~\cite{ref09}. The main contributions of this study are as follows.

\begin{itemize}

\item This paper proposes a supervised learning-based shadow removal model that combines image inpainting and knowledge distillation techniques. To achieve this, we
present two image-inpaining-based image generation model architectures and loss functions for efficient learning.

\item We trained our model using the correlations between pixels in the non-shadow regions of the shadowed and shadow-removed images. By incorporating this information, we correct the pixel values for the entire shadow-free image. This post-processing operation can help restore the overall colors of the generated images to closely match those of the original images and enhance shadow removal performance.

\item We conducted extensive experiments using the ISTD+dataset, which is a public dataset, and demonstrated that the proposed SFR-Net can achieve a performance similar to that of the state-of-the-art methods.

\end{itemize}
\section{Related Works}
\label{sec:related}

Previous studies on shadow removal primarily used physics-based methods~\cite{ref25,ref26,ref27}. Physics-based methods remove shadows based on prior environmental information, including lighting and space. Since the inception of deep learning, many shadow-removal techniques based on generative models, such as generative adversarial networks (GANs)~\cite{ref28} and diffusion models~\cite{ref29}, have been proposed in recent years~\cite{ref01,ref02,ref08,ref09,ref10,ref11}. Qu et al.~\cite{ref17} proposed a multicontext model that extracts shadow-matte images. After using a global localization network (G-Net), appearance modeling network (A-Net), and semantic modeling network (S-Net) to extract context, matte images were generated through a $1 \times 1$ convolution to remove shadows. Cun et al.~\cite{ref18} used a hierarchical attention aggregation node that utilized an attention mechanism to hierarchically aggregate multiple contexts. They designed two nodes to generate shadow-free images and masks. In addition, they used a shadow-matting GAN to generate a shadow matte to augment and use the shadow image. Le et al.~\cite{ref08} interpreted shadow removal from the perspective of image decomposition. They proposed SP+M Net, which extracts a reliable image based on a physical lighting model for images (SP-Net) and then generates a shadow matte image (M-Net) by utilizing it. Moreover, in a follow-up study, they introduced SP+M+I Net, which enhanced the overall performance by applying an image inpainting network (I-Net) to generate shadow-removed images~\cite{ref02}. They also released the ISTD+ dataset, which was generated by performing color restoration preprocessing on the GT to transform the nonshadowed regions of the GT images to be highly similar to those of the shadowed images.

From a model architectural perspective, recent shadow-removal methods have increasingly adopted transformer- and diffusion-based learning frameworks. HomoFormer~\cite{ref30} employed a homogenized transformer architecture to efficiently capture long-range dependencies in shadowed regions. CRFormer~\cite{ref19} introduced region-aware cross attention between shadow and non-shadow patches and applied RefineNet to recover information lost during the shadow removal process. Guo et al.~\cite{ref10} proposed a model that leverages both spatial and channel-wise information using shadow interaction and channel attention modules. ShadowGAN-Former~\cite{ref31} reweighted self-attention mechanisms using shadow masks to better model both the global context and local shadow structures within a GAN-transformer hybrid architecture.

Simultaneously, diffusion-based approaches have gained attention because of their ability to generate high-quality shadow-free images. DeS3~\cite{ref09} used a pretrained DINO-ViT~\cite{ref32} to extract features during the reverse diffusion process and incorporated them into the loss function. Similarly, ShadowDiffusion~\cite{ref33} integrated diffusion models with degradation priors to enhance restoration performance. Mei et al.~\cite{ref34} proposed a latent feature-guided diffusion model that learns shadow-free priors using a two-stage training approach to address the challenge of texture preservation in shadowed regions. Another recent method, Inpaint4Shadow~\cite{ref35}, adapts image inpainting techniques specifically for shadow removal and demonstrates competitive results in both shadow elimination and texture consistency.

The main limitation of most previous studies is that they could not address the differences in color between shadowed and GT images. Regarding human perception, the sense of incongruity can be greater when shadows are removed from images with simple colors and textures than when they are removed from complex images. This problem is accentuated by the difference in color between the shadow image and the GT image when the shadow removal model is trained from scratch. This issue may be attributed to both the limitations related to colors and the fact that the boundary of the shadowed region was unclear when the shadow was removed. Consequently, the shadow was not properly removed, resulting in a greater sense of incongruity. Therefore, if a shadow removal model is trained without considering these factors, it may generate images in which the boundary between the shadowed and non-shadowed regions is unclear, or there is a loss of color.

The proposed approach incorporates feature-level knowledge distillation between the shadow image and the GT image to minimize the degradation of the shadow region during the shadow removal process and trains the model to generate shadow-removed images that look more natural and resemble GT images. Moreover, our approach can restore images with minimal loss to the non-shadow region of the original shadow images by utilizing a pretrained image inpainting model in the shadow removal process.

\begin{figure}[!t]
    \centering
    \includegraphics[width=\linewidth]{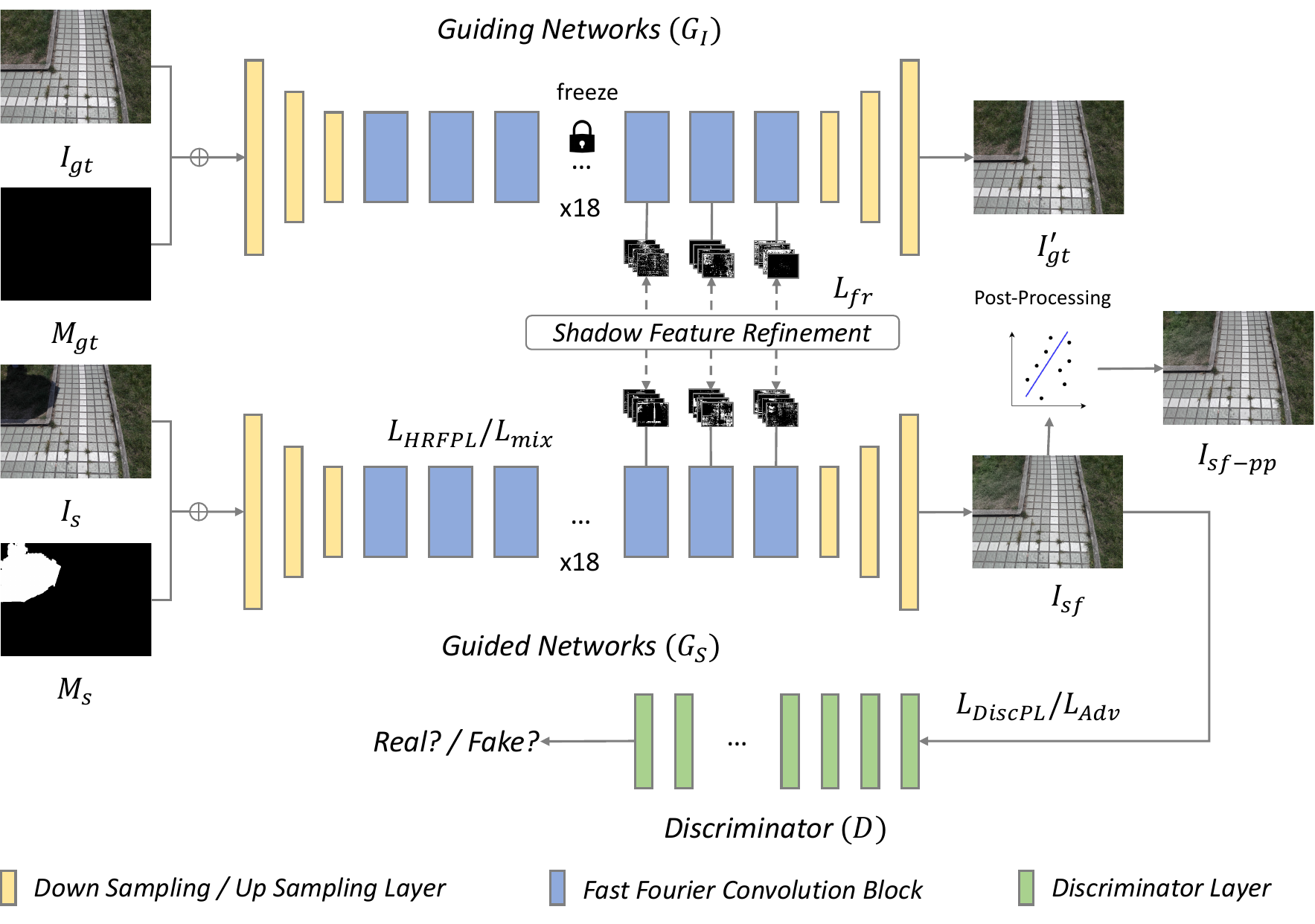}
    \caption{Network architecture of the proposed shadow feature refinement network.}
    \label{fig:fig1}
\end{figure}
\section{Methodology}
\label{sec:method}

This section introduces the proposed shadow removal model, referred to as SFR-Net. The SFR-Net utilizes a deep-learning-based generative model to generate a shadow-removed image, $I_{sf}$, using the shadow image $I_S$ and binary mask $M_S$ for the shadow region of the shadow image, which are provided as inputs. The shadow mask $M_S$ was obtained from the ground-truth mask annotations provided in the ISTD+ dataset during the training and testing phases. For datasets without official mask annotations, such as the shadow removal dataset (SRD), commonly used predicted masks from models such as DHAN~\cite{ref18} were used as input masks. Subsequently, the designed postprocessing was applied to the shadow-removed image to generate the final shadow-removed image $I_{sf\text{-}pp}$, wherein the colors were restored. Figure~\ref{fig:fig1} shows the structure of the SFR-Net.

\begin{figure*}[!t]
    \centering
    \includegraphics[width=\linewidth]{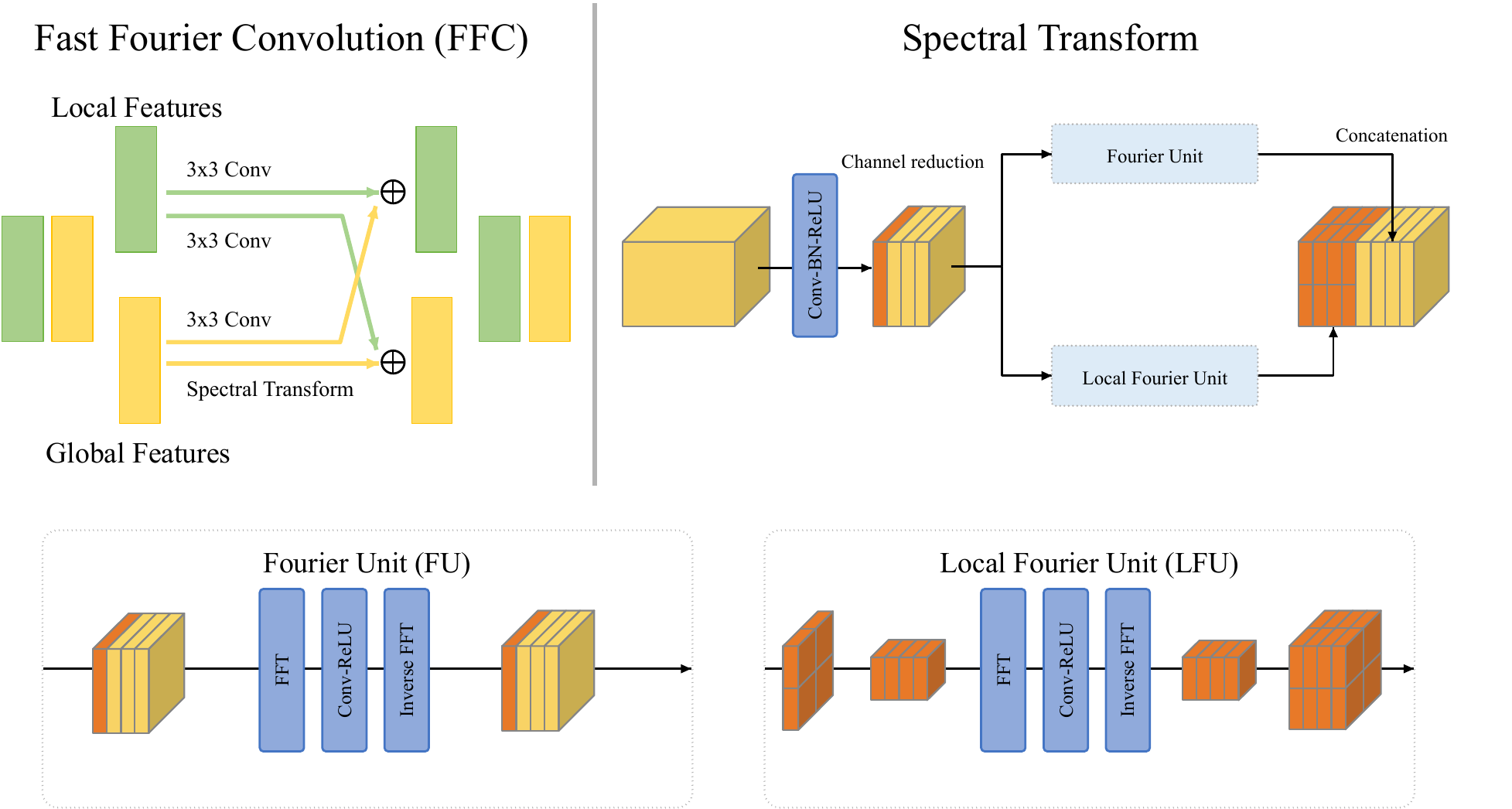}
    \caption{Architecture of FFC block and spectral transform.}
    \label{fig:fig2}
\end{figure*}

\subsection{Shadow Feature Refinement Networks}
\label{subsec:networks}

The SFR-Net is a generative adversarial network (GAN)-based generative model that consists of two image generators and one discriminator. As mentioned previously, the two generators were designed based on the hypothesis that more realistic images can be generated by comparing the differences between the GT and predicted images, as well as the features of the intermediate layers. These two generators are referred to as the guiding network $G_I$ and guided network $G_S$. The $G_S$ uses the shadow image $I_S$ and paired mask image $M_S$ as inputs to remove shadows. Simultaneously, the pretrained $G_I$ takes $I_{gt}$, which is the GT image for $I_S$, and the empty mask image $M_{gt}$ as input to reconstruct $I'_{gt}$. The $G_S$ compares features generated in the intermediate layers during the process of shadow removal with features in the intermediate layers of the $G_I$ while proceeding with learning. In this study, we used LaMa~\cite{ref36} for $G_I$ and $G_S$ among the various image inpainting models because it can learn rich contexts using fast Fourier convolution (FFC)~\cite{ref37}.

As shown in Figure~\ref{fig:fig2}, the fast Fourier convolution (FFC) module applies a spectral transform to process global and local features separately and efficiently in parallel. This design enables the network to learn rich contextual representations essential for complex tasks such as shadow removal. The spectral transform is composed of two core components: the Fourier unit (FU) and local Fourier unit (LFU). FU processes global features by applying a fast Fourier transform (FFT) to project spatial features into the frequency domain, where long-range dependencies and global structures can be captured more effectively. By contrast, LFU focuses on enhancing the local detail by dividing the feature map into four quadrants, processing each separately with FFT, and fusing them via channel-wise concatenation. It further applies spatial shift operations to strengthen the interactions between neighboring regions.

These local features are then fused with global features, thereby allowing the model to preserve both broad contextual relationships and fine-grained details simultaneously. Figure~\ref{fig:fig2} illustrates this architecture, including the overall structure of the FFC block and the internal workflows of the FU and LFU. The proposed guided network $G_S$ is constructed using an FFC-based architecture. Specifically, the $G_S$ consists of four DownScale layers based on FFC, two FFC layers, 18 FFCResNetBlocks using skip connections, and four UpScale layers, which are also based on FFC. In addition, the discriminator uses the patch-based discriminator used in the Pix2PixHD model~\cite{ref38}. The shadow-removed image $I_{sf}$, which is generated by the guided network $G_S$, is subjected to a post-processing algorithm that restores the overall colors degraded during the shadow removal process and becomes the final shadow-free image $I_{sf\text{-}pp}$. The proposed postprocessing algorithm learns the changes in the pixel values of the nonshadowed region between $I_{sf}$ and $I_S$. Subsequently, it refines the pixel values for the entire region of $I_{sf}$, including the shadowed region. Through this process, the overall color of $I_{sf}$ can be restored to resemble that of the input image.

\subsection{Loss Function}
\label{subsec:loss}

The SFR-Net uses adversarial loss ($L_{Adv}$) for generator $G_S$ and discriminator $D$, high receptive field perceptual loss ($L_{HRFPL}$)~\cite{ref36}, which is a ResNet-based perceptual loss, and discriminator feature matching loss ($L_{fm}$)~\cite{ref39}. In addition, to improve the image reconstruction performance, SFR-Net uses mix loss ($L_{mix}$)~\cite{ref40}, which combines MS-SSIM~\cite{ref41} loss and $L_1$ loss as the reconstruction loss function. Finally, it uses feature refinement loss ($L_{fr}$), which uses knowledge distillation between the guided network ($G_S$) and guiding network ($G_I$) to improve the shadow removal performance. The overall joint loss function used is defined as shown in Equation~\eqref{eq:total_loss}:

\begin{small}
\begin{equation}
\label{eq:total_loss}
\begin{aligned}
L &= \alpha L_{Adv} + \beta L_{HRFPL} + \gamma L_{fm}\\
  &\quad + \delta L_{mix} + \lambda_{fr} L_{fr} + \zeta R_1,
\end{aligned}
\end{equation}
\end{small}

where $R_1$ denotes the gradient penalty, and $\alpha$, $\beta$, $\gamma$, $\delta$, $\lambda_{fr}$, and $\zeta$ denote weight values for each loss.

\paragraph{Adversarial loss.}
SFR-Net is a type of GAN that uses an adversarial loss function. It further calculates the loss function separately for the shadowed and non-shadowed regions. The non-shadow region of the image generated by $G_S$ of the SFR-Net (it is the inverted mask region and is referred to as $1-M_S$) should be identical to that of the shadow-free image ($I_{gt}$). Hence, the discriminator is trained to classify the generated non-shadowed region as a real image. Therefore, it is preferable for the pixel values in the non-shadow region to remain unchanged. The discriminator is trained to assign high confidence scores to these regions, as they should match the target (i.e., ground-truth) distribution. By contrast, the shadow region is expected to be altered by the generator, and the discriminator is trained to detect these changes accordingly. In our patch-based discriminator approach, each image is divided into $N$ spatial patches, and the loss is computed on a per-patch basis. Each patch is supervised differently depending on whether it belongs to the shadow region, as indicated by the shadow mask. Specifically, patches within the shadow regions are expected to be modified by the generator and should be distinguished from the ground truth, whereas patches outside the shadow region should be consistent with the original image content. The discriminator loss ($L_D$) is defined as follows:

\begin{equation}
\label{eq:discriminator_loss}
\begin{aligned}
L_D = &\frac{1}{N}\sum_{i=1}^{N}\Bigl[ -\log\bigl(D_i(I_{gt})\bigr) \\
&-M_{S,i}\log\bigl(1-D_i(I_{sf})\bigr) \\ 
&-(1-M_{S,i})\log\bigl(D_i(I_{sf})\bigr)\Bigr].
\end{aligned}
\end{equation}

Here, $D_i(\cdot)$ represents the discriminator output for the $i$-th patch, and $M_{S,i}\in\{0,1\}$ indicates whether the patch belongs to a shadow region. Specifically, when $M_{S,i}=1$ (shadow region), the corresponding patch is supervised to reflect the transformation expected in the shadow regions, whereas patches with $M_{S,i}=0$ (non-shadow region) are encouraged to retain consistency with the target image. This resulted in a conditional loss formulation that applied region-specific supervision to each patch. Because the SFR-Net adopts a patch-based discriminator, the overall loss is computed over spatially divided regions of the input image~\cite{ref38}. Equation~\eqref{eq:generator_loss} expresses the generator loss ($L_G$) mathematically and allows the discriminator to classify it as a real image based on the discriminator output for $I_{sf}$. Equation~\eqref{eq:adversarial_loss} represents the final adversarial loss. We applied a nonsaturating adversarial loss~\cite{ref42} that can minimize $L_D$ and $L_G$ by applying a stop gradient to the losses in Equations~\eqref{eq:discriminator_loss} and~\eqref{eq:generator_loss} for the adversarial loss:

\begin{small}
\begin{equation}
\label{eq:generator_loss}
L_G = -\mathbb{E}\left[\log\bigl(D(I_{sf})\bigr)\right].
\end{equation}
\end{small}

\begin{small}
\begin{equation}
\label{eq:adversarial_loss}
L_{Adv} = \operatorname{sg}_{\theta}(L_D)
+ \operatorname{sg}_{\xi}(L_G)
\longrightarrow \min_{\theta,\xi}.
\end{equation}
\end{small}

\paragraph{High receptive field perceptual loss.}
It is challenging to generate high-quality images using simple pixel-wise evaluations when generating images in image-to-image translation tasks~\cite{ref43}. The calculated pixel distances between images often converge to an average value, resulting in blurred images. Several studies have used perceptual loss to address this issue~\cite{ref36,ref38,ref43,ref44}. Perceptual loss can effectively preserve the overall color and texture of images, thereby enhancing image quality. This was achieved by comparing the high-level perceptual qualities of images using a trained network. SFR-Net employs a high receptive field perceptual loss ($L_{HRFPL}$) to understand the global context~\cite{ref36}. $L_{HRFPL}$ has demonstrated excellent performance in high-resolution image-restoration tasks such as image inpainting and shadow removal using ResNet-based perceptual loss~\cite{ref36}.

\begin{small}
\begin{equation}
\label{eq:hrfpl}
L_{HRFPL} = \sum_{j\in\mathcal{L}}\frac{1}{C_jH_jW_j} ||\varphi_{HRF_j}(I_{gt})-\varphi_{HRF_j}(I_{sf})||^2.
\end{equation}
\end{small}

Here, $L_{HRFPL}$ calculates the perceptual loss for each layer included in $\mathcal{L}$, which is a set of target layers, as detailed in Equation~\eqref{eq:hrfpl}. $\varphi_{HRF_j}(\cdot)$ represents the output of the activation function for the $j$-th layer of the ResNet-based pretrained model implemented using Fourier convolution or dilated convolution. In addition, $C_j$, $H_j$, and $W_j$ denote the channel, height, and width of the $j$-th layer of the feature map, respectively.

\paragraph{Discriminator feature matching loss.}
Perceptual loss assists in training by comparing the perceptual quality, ensuring that the distribution of the generated images is similar to that of the real images. However, it may not be suitable for all image-to-image translation fields, such as shadow removal, because it uses a pretrained model that has already been trained for classification. The discriminator feature matching loss ($L_{fm}$) can be optimized to compare the perceptual quality of each generative task. Hence, the discriminator feature matching loss and the perceptual loss~\cite{ref39} serve to complement each other. Furthermore, $L_{fm}$ compares the outputs of the activation functions of the intermediate layers of the discriminator for both $I_{gt}$ and $I_{sf}$. $L_{fm}$ is defined as follows:

\begin{small}
\begin{equation}
\label{eq:feature_matching}
L_{fm} = \sum_{j\in\mathcal{L}}\frac{1}{C_jH_jW_j} \lVert D_j(I_{gt})-D_j(I_{sf})\rVert^2.
\end{equation}
\end{small}

Here, $\mathcal{L}$ is a set of the target layers, and $D_j(\cdot)$ represents the output of the activation function for the $j$-th layer of the discriminator.

\paragraph{Mix loss.}
In image-to-image translation tasks, many models employ the $L_1$ loss for image reconstruction. However, $L_1$ loss compares pixel-wise differences and does not align with the mechanisms of the human visual system (HVS), which is more sensitive to luminance, contrast, and structural information. Therefore, the structural similarity index measure (SSIM), which can quantitatively evaluate these factors, is suitable for reconstruction loss~\cite{ref40}. Consequently, we used mix loss ($L_{mix}$)~\cite{ref40}, which combines $L_1$ and the multi-scale structural similarity index measure (MS-SSIM).

\begin{equation}
\label{eq:ssim_components}
\begin{aligned}
l(p) &= \frac{2\mu_x\mu_y+C_1}{\mu_x^2+\mu_y^2+C_1}, \\
c(p) &= \frac{2\sigma_x\sigma_y+C_2}{\sigma_x^2+\sigma_y^2+C_2}, \\
s(p) &= \frac{\sigma_{xy}+C_3}{\sigma_x\sigma_y+C_3}.
\end{aligned}
\end{equation}

The SSIM calculates the luminance, contrast, and structure for each pixel position $p$ of the input image $x$ and comparison image $y$ as $l(p)$, $c(p)$, and $s(p)$, as shown in Equation~\eqref{eq:ssim_components}. $\mu$ and $\sigma$ denote the mean and standard deviation of the images, respectively, and $C_1$, $C_2$, and $C_3$, which are used in each equation, are variables that can be set. Here, $C_n=(K_nL)^2$, and a normalized value of 1 was used as the pixel range $L$ (the RGB pixel range was 0--255). $K_1=0.01$ is used as the constant for $C_1$, and $K_2=0.03$ is used as the constant for $C_2$. For $C_3$, $C_3=C_2/2$ is used to simplify the mathematical expression. SSIM values were calculated by multiplying $l(p)$, $c(p)$, and $s(p)$. By substituting $C_3=C_2/2$, Equation~\eqref{eq:ssim} is obtained.

\begin{equation}
\label{eq:ssim}
\begin{aligned}
\operatorname{SSIM}(p)
&= \frac{2\mu_x\mu_y+C_1}{\mu_x^2+\mu_y^2+C_1}
\cdot\frac{2\sigma_{xy}+C_2}{\sigma_x^2+\sigma_y^2+C_2} \\
&= l(p)\cdot cs(p).
\end{aligned}
\end{equation}

To eliminate the dependency on the mean and standard deviation when using SSIM as the loss function, we calculated these statistics for each patch $P$ using a convolution operation. This filter employed a Gaussian filter $G_{\sigma_G}$, with a standard deviation of $\sigma_G$ as its weight. The convolution filter size was equal to the patch size, and a size of $35\times35$ pixels was used. Moreover, because a convolution filter is used, the SSIM of patch $P$ can be calculated through operations on $\tilde{p}$, which is the central pixel of the patch. However, the SSIM can affect the image quality depending on the hyperparameter value $\sigma_G$ of $G_{\sigma_G}$. To mitigate this sensitivity, the MS-SSIM can be calculated using a Gaussian filter with multi-scale $\sigma_G$, as shown in Equation~\eqref{eq:ms_ssim}.

\begin{equation}
\label{eq:ms_ssim}
\operatorname{MS\text{-}SSIM}(P)
= l_M^{\alpha}(\tilde{p})\cdot\prod_{j=1}^{M}cs_j^{\beta_j}(\tilde{p}).
\end{equation}

The values defined in Equation~\eqref{eq:ssim} for the scales $M$ and $j$ were used for $l_M$ and $cs_j$, respectively. The parameters $\alpha=\beta_j=1$ are used, $j=\{1,\ldots,M\}$ is the set for the order of the scale, and $\sigma_G=[0.5,1.0,2.0,4.0,8.0]$ is adopted as the standard deviation of the scale. $l(\tilde{p})$ uses the value when $\sigma_G=8.0$, whereas $cs(\tilde{p})$ uses the value that is computed by applying a Gaussian filter to each scale with standard deviations $\sigma_G=[0.5,1.0,2.0,4.0,8.0]$ and then multiplying the results. In this case, the hyperparameters were adopted directly as originally proposed by Zhao et al., without any modifications. As shown in Equation~\eqref{eq:ms_ssim_loss}, the equation is transformed to use the MS-SSIM as the loss.

\begin{equation}
\label{eq:ms_ssim_loss}
\begin{aligned}
L_{ms\text{-}ssim}(P)
&=1-\operatorname{MS\text{-}SSIM}(\tilde{p}) \\
&=1-l_M^{\alpha}(\tilde{p})\cdot\prod_{j=1}^{M}cs_j^{\beta_j}(\tilde{p}).
\end{aligned}
\end{equation}

Finally, Equation~\eqref{eq:mix_loss} shows the $L_{mix}$ loss, which is a combination of $L_{ms\text{-}ssim}$ and $L_1$ losses, expressed as follows:

\begin{equation}
\label{eq:mix_loss}
L_{mix}=\gamma_{mix}\cdot L_{ms\text{-}ssim}
+(1-\gamma_{mix})\cdot G_{\sigma_G^{M}}\cdot L_1.
\end{equation}

The mixing coefficient $\gamma_{mix}$ was experimentally set to 0.025 and is distinct from the overall loss weight $\gamma$ in Equation~\eqref{eq:total_loss}. $L_1$ is multiplied by the Gaussian filter $G_{\sigma_G^{M}}$ for scale $M$, based on $\tilde{p}$ for each patch. This is because $L_{ms\text{-}ssim}$ calculates the loss values for the pixels surrounding the center pixel $\tilde{p}$.

\begin{figure}[!t]
    \centering
    \includegraphics[width=\linewidth]{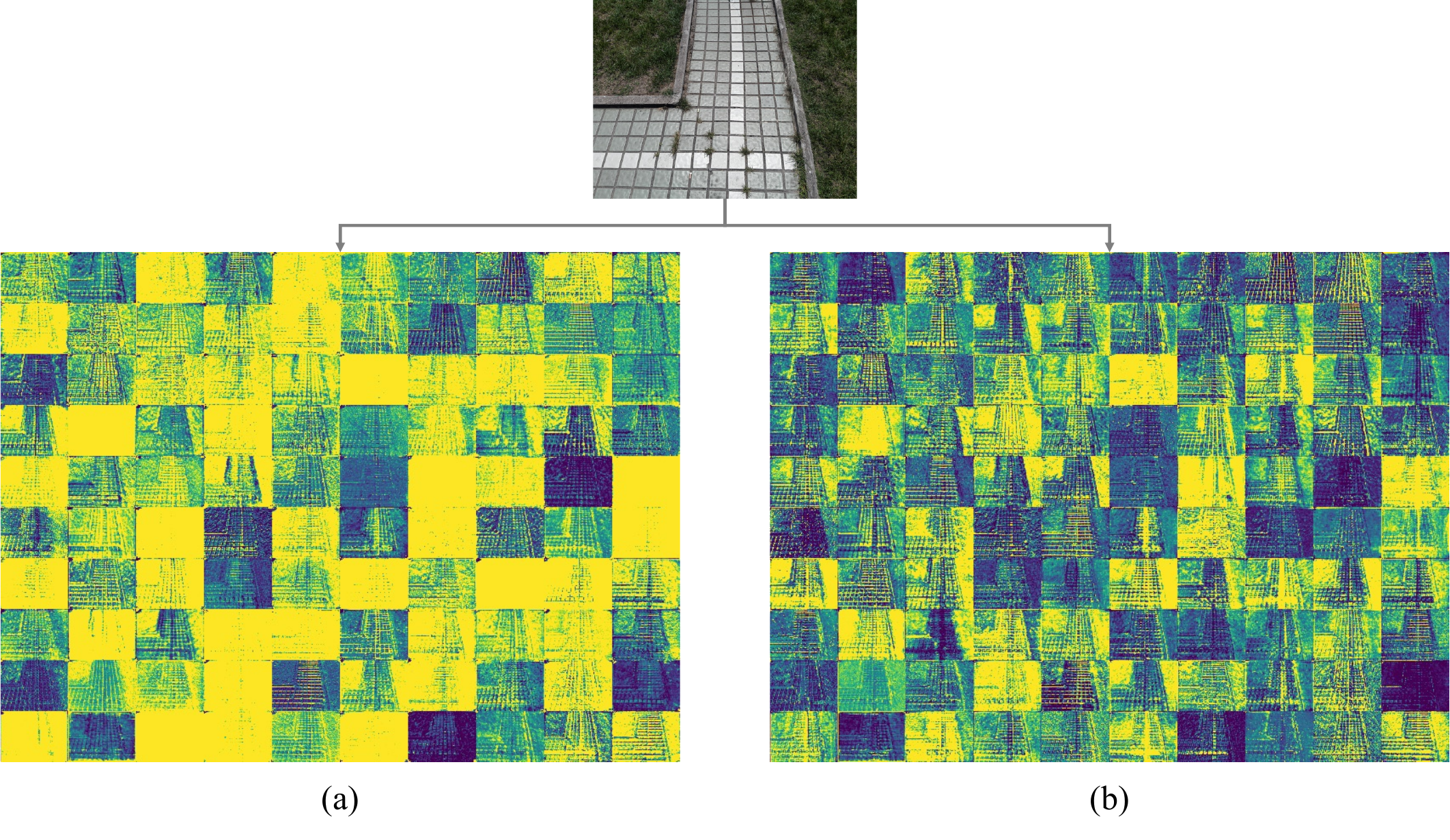}
    \caption{Comparison of SFR-Net’s shadow-removed feature maps. (a) Without $L_{fr}$, (b) With $L_{fr}$.}
    \label{fig:fig3}
\end{figure}

\paragraph{Feature refinement loss.}
In general, perceptual loss is used to supplement pixel-level losses such as $L_1$ and $L_2$ losses in image translation tasks. It also enables the comparison of high-level perceptual losses using a pretrained model~\cite{ref43}. The feature matching loss was developed based on the hypothesis that a discriminator, when trained on the same dataset as the generative model, can calculate perceptual losses more effectively because of its enhanced ability to extract features from the same data domain~\cite{ref39}. Similarly, we assumed that if the image inpainting model, which is similar to the shadow removal task, is trained such that the features extracted through feature refinement loss ($L_{fr}$) when reconstructing GT images are similar to the features extracted when removing shadows, the perceptual loss will be reduced, and images that look more natural will be produced. Figure~\ref{fig:fig3} shows a comparison of the feature maps from our proposed model (20th FFC Block) with and without the application of the proposed $L_{fr}$. The results demonstrate that the feature map contained more diverse and richer information when the proposed $L_{fr}$ algorithm was applied. As shown in Figure~\ref{fig:fig1}, SFR-Net uses $I_{gt}$ and its corresponding empty mask $M_{gt}$ as inputs and extracts the feature map from the pretrained $G_I$. Furthermore, $G_S$ learns from the feature map extracted by $G_I$ via feature distillation. In other words, $G_S$ learns the image representations of $I_{gt}$ reconstructed by $G_I$ and refines the features. Here, the features of $G_S$ and $G_I$ are compared using $L_2$ distance and used as the feature refinement loss ($L_{fr}$) function. The equation for $L_{fr}$ is defined in Equation~\eqref{eq:feature_refinement}.

\begin{small}
\begin{equation}
\begin{aligned}
\label{eq:feature_refinement}
L_{fr}=\frac{1}{N}\sum_{l\in\mathcal{L}}\frac{1}{C_lH_lW_l}
||G_I^l(I_{gt}\oplus M_{gt}) \\ 
-G_S^l(I_S\oplus M_S)||^2.
\end{aligned}
\end{equation}
\end{small}

Here, $\mathcal{L}$ is a set of target layers that perform knowledge distillation of $G_S$ and $G_I$, and $N$ is the total number of layers included in $\mathcal{L}$. $G_I^l$ and $G_S^l$ are the outputs of the $l$-th layer's activation functions for $G_I$ and $G_S$, respectively. The input values for $G_I$ and $G_S$ are four-dimensional images, $(I_{gt}\oplus M_{gt})$ and $(I_S\oplus M_S)$, obtained by combining three-dimensional images with one-dimensional mask images. $\oplus$ denotes channel concatenation for the images. The model successfully refines the features using $L_{fr}$ for shadow removal, which is a downstream task in image inpainting.

\subsection{Post-processing for Color Restoration}
\label{subsec:postprocessing}

SFR-Net combines five different loss functions to generate natural shadow-removed images. However, the main limitation is that the overall color of the generated images can be altered. This problem occurs because, in most shadow removal datasets, the data pair of a shadow image and a GT image exhibits a difference in color between the two images owing to differences in conditions, such as when the image was captured and lighting conditions~\cite{ref23}. As the training of the shadow removal model progresses, it discovers and learns changes in shadows as well as changes in colors owing to lighting conditions between the shadow and GT images. Consequently, the overall color of the images was altered. To overcome this problem, Le et al. incorporated a linear regression algorithm to create an ISTD+ dataset with a reduced color loss~\cite{ref08}. By contrast, we incorporated a post-processing algorithm to restore the colors of the shadow-removed image to match those of the input image. Here, $I_S$ is an image in which shadows are cast on $I_{gt}$. Therefore, based on the assumption that $I_S$ and $I_{gt}$ have no color or pixel differences in the non-shadowed region, the shadow-free image $I_{sf}$ should have no color differences in the non-shadowed region. Therefore, we learned a model (e.g., regression) for color differences in the non-shadowed region between $I_S$ and $I_{sf}$ and applied it to the entire area of $I_{sf}$ to restore the altered colors in $I_{sf}$. Various methods can be used to determine these color differences. For example, if linear regression analysis is applied, the equation can be represented as follows:

\begin{equation}
\label{eq:color_regression}
I_S\odot(1-M_S)=\beta_0+\beta_1\cdot\bigl(I_{sf}\odot(1-M_S)\bigr).
\end{equation}

\begin{equation}
\label{eq:color_restoration}
I_{sf\text{-}pp}=\beta_0+\beta_1\cdot I_{sf}.
\end{equation}

As shown in Equation~\eqref{eq:color_regression}, the inverted mask $(1-M_S)$ is used to identify the non-shadow regions in $I_S$ and $I_{sf}$ and derive the linear regression relationship between them. Further, $\beta_0$ and $\beta_1$ denote the intercept and regression coefficient (slope) for the linear regression, respectively. $\beta_0$ and $\beta_1$ are applied to all pixels of $I_{sf}$ through linear regression trained for each image via Equation~\eqref{eq:color_restoration} to calculate the final output ($I_{sf\text{-}pp}$). Here, $\odot$ denotes element-wise multiplication. The prediction of each pixel was performed separately for R, G, and B.

\section{Experiments}
\label{sec:experiments}

\subsection{Experimental Setup}
\label{subsec:experimental_setup}

\noindent\textbf{Datasets.} The performance of the proposed method was verified using two publicly available shadow-removal datasets, ISTD+ and SRD. The ISTD+ dataset is a shadow-removal dataset obtained by postprocessing and adjusting the ISTD dataset. This dataset resulted in a slight improvement in the color differences between shadowed and shadow-free images~\cite{ref08}. The ISTD+ dataset is a large-scale shadow-removal dataset consisting of 1,870 triplets, where each triplet is composed of a shadow image, a shadow-free image, and a mask image. The resolution of the images was $640 \times 480$, and the dataset consists of 1,330 triplets for the training set and 540 triplets for the test set~\cite{ref45}. The SRD dataset comprises 3,088 pairs of shadowed and shadow-free images; however, it does not provide ground-truth mask annotations and has no fixed resolution. To ensure a fair comparison with ISTD+, we resized all the SRD images to $640 \times 480$ pixels. Moreover, because our method and the baseline models require triplet data, including shadow masks, we adopted the widely used approach of generating masks using the DHAN model~\cite{ref18} for both the training and evaluation of the SRD.

\noindent\textbf{Metrics.} As in previous studies, we quantitatively measured the root mean square error (RMSE) for the GT and shadow-free images that were transformed into the LAB color space. We also measured the peak signal-to-noise ratio (PSNR) and structural similarity index measure (SSIM) for RGB space. Each of these metrics is evaluated for shadowed, non-shadowed, and entire regions. Smaller RMSE values are ideal, whereas higher PSNR and SSIM values indicate a better image quality. The equations for each evaluation metric are as follows:
\begin{equation}
\mathrm{RMSE} = \sqrt{\frac{1}{N}\sum_{i=1}^{N}(x_i-y_i)^2}.
\label{eq:rmse}
\end{equation}
\begin{equation}
\mathrm{PSNR} = 10\log\left(\frac{N\,255^2}{\sum_{i=1}^{N}(x_i-y_i)^2}\right).
\label{eq:psnr}
\end{equation}
\begin{equation}
\mathrm{SSIM} = \frac{2\mu_x\mu_y+C_1}{\mu_x^2+\mu_y^2+C_1}
\cdot\frac{2\sigma_{xy}+C_2}{\sigma_x^2+\sigma_y^2+C_2}.
\label{eq:ssim_metric}
\end{equation}
Here, $x$ and $y$ represent the input and comparison images, respectively. In addition, $N$ denotes the total number of pixels in the image, and $x_i$ and $y_i$ represent the $i$-th pixels in the $x$ and $y$ images, respectively.

\noindent\textbf{Implementation details.} The proposed model was developed using PyTorch 2.0 (CUDA 11.8) and PyTorch Lightning 1.2.9, and was trained using an NVIDIA Tesla A100 GPU. SFR-Net utilizes LaMa~\cite{ref36}, which possesses state-of-the-art (SOTA) performance, as the base model for knowledge distillation using an image inpainting model. The model used had four downsampling layers, 18 FFC residual blocks, and four upsampling layers. We used the Pix2PixHD-based discriminator proposed by Wang et al.~\cite{ref38}. We set the batch size to four and used Adam and Lion optimizers. The learning rate was set to 0.001 for the generator and 0.0001 for the discriminator. The mixing coefficient in Equation~\eqref{eq:mix_loss} was set to $\gamma_{mix}=0.025$, consistent with Section~\ref{subsec:loss}.

The weight $\lambda_{fr}$, which is multiplied by $L_{fr}$, varies for each model. The experimental outcomes indicated that the best performance was achieved when $\lambda_{fr}=100$. Similarly, in Equation~\eqref{eq:total_loss}, we used $\alpha=10$, $\beta=30$, $\gamma=100$, $\delta=10$, and $\zeta=0.001$ as the weight values that are multiplied by each loss. These values were selected based on the experiments because they yielded the best performance.

\begin{figure*}[!t]
    \centering
    \begin{subfigure}[b]{0.12\textwidth}
        \centering
        \includegraphics[width=\linewidth,viewport=0 0 137 537,clip]{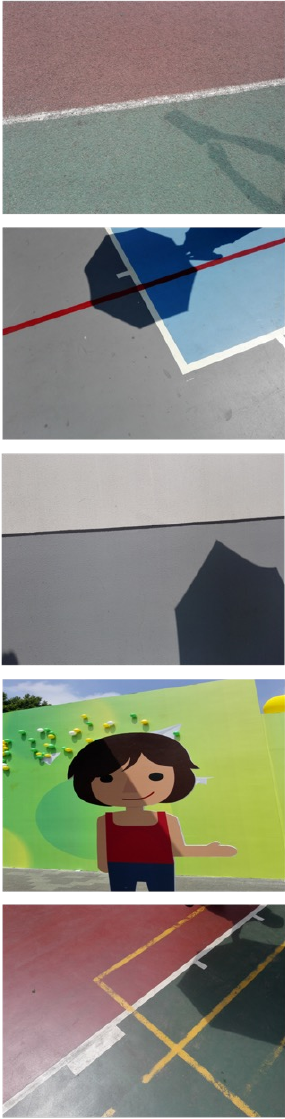}
        \caption{}\label{fig:fig4a}
    \end{subfigure}\hfill
    \begin{subfigure}[b]{0.12\textwidth}
        \centering
        \includegraphics[width=\linewidth,viewport=144 0 281 537,clip]{figures/fig4.pdf}
        \caption{}\label{fig:fig4b}
    \end{subfigure}\hfill
    \begin{subfigure}[b]{0.12\textwidth}
        \centering
        \includegraphics[width=\linewidth,viewport=286 0 423 537,clip]{figures/fig4.pdf}
        \caption{}\label{fig:fig4c}
    \end{subfigure}\hfill
    \begin{subfigure}[b]{0.12\textwidth}
        \centering
        \includegraphics[width=\linewidth,viewport=428 0 565 537,clip]{figures/fig4.pdf}
        \caption{}\label{fig:fig4d}
    \end{subfigure}\hfill
    \begin{subfigure}[b]{0.12\textwidth}
        \centering
        \includegraphics[width=\linewidth,viewport=571 0 708 537,clip]{figures/fig4.pdf}
        \caption{}\label{fig:fig4e}
    \end{subfigure}\hfill
    \begin{subfigure}[b]{0.12\textwidth}
        \centering
        \includegraphics[width=\linewidth,viewport=713 0 850 537,clip]{figures/fig4.pdf}
        \caption{}\label{fig:fig4f}
    \end{subfigure}\hfill
    \begin{subfigure}[b]{0.12\textwidth}
        \centering
        \includegraphics[width=\linewidth,viewport=856 0 993 537,clip]{figures/fig4.pdf}
        \caption{}\label{fig:fig4g}
    \end{subfigure}\hfill
    \begin{subfigure}[b]{0.12\textwidth}
        \centering
        \includegraphics[width=\linewidth,viewport=997 0 1134 537,clip]{figures/fig4.pdf}
        \caption{}\label{fig:fig4h}
    \end{subfigure}

    \caption{Qualitative evaluation results of SFR-Net and state-of-the-art
    shadow removal methods for the ISTD+ dataset. (a) Input,
    (b) ShadowDiffusion~\cite{ref33}, (c) HomoFormer~\cite{ref30},
    (d) ShadowFormer~\cite{ref10}, (e) BMNet~\cite{ref01},
    (f) Ours, (g) InstanceShadowDiffusion~\cite{ref34}, (h) GT.}
    \label{fig:fig4}
\end{figure*}
\begin{table*}[t]
\centering
\caption{Quantitative evaluation results of SFR-Net and state-of-the-art shadow removal methods on the ISTD+ dataset.}
\label{tab:tab1}
\begingroup
\setlength{\tabcolsep}{4pt}
\renewcommand{\arraystretch}{1.15}
\small
\resizebox{\textwidth}{!}{%
\begin{tabular}{lccccccccc}
\toprule
\multicolumn{1}{c}{\multirow{2}{*}[-0.8ex]{Method}} & \multicolumn{3}{c}{Shadow Region} & \multicolumn{3}{c}{Non-Shadow Region} & \multicolumn{3}{c}{All} \\
\cmidrule(lr){2-4} \cmidrule(lr){5-7} \cmidrule(lr){8-10}
 & RMSE$\downarrow$ & SSIM$\uparrow$ & PSNR$\uparrow$ & RMSE$\downarrow$ & SSIM$\uparrow$ & PSNR$\uparrow$ & RMSE$\downarrow$ & SSIM$\uparrow$ & PSNR$\uparrow$ \\
\midrule
Shadow Image & 30.4831 & 0.9265 & 20.8095 & 3.1493 & 0.9552 & 33.8855 & 12.4080 & 0.8747 & 20.2209 \\
ShadowDiffusion~\cite{ref33} & 5.8667 & 0.9782 & 35.8192 & 3.2892 & 0.9253 & 32.8163 & 3.8768 & 0.8995 & 30.3881 \\
HomoFormer~\cite{ref30} & \underline{5.0098} & 0.9840 & 37.4862 & 3.1282 & 0.9350 & 33.3300 & 3.5245 & 0.9161 & 31.3985 \\
Inpaint4Shadow~\cite{ref35} & 6.5203 & 0.9621 & 35.2136 & 4.4683 & 0.8234 & 29.8008 & 4.9124 & 0.7783 & 28.2474 \\
ShadowFormer~\cite{ref10} & 5.0755 & \underline{0.9864} & \textbf{38.0692} & \textbf{2.8663} & \underline{0.9550} & \textbf{35.1463} & \textbf{3.3461} & \underline{0.9387} & \textbf{32.7827} \\
BMNet~\cite{ref01} & 5.5477 & \textbf{0.9865} & 36.8070 & \underline{2.9791} & \textbf{0.9576} & 34.4725 & 3.5375 & \textbf{0.9410} & 31.8521 \\
SG-ShadowNet~\cite{ref11} & 6.0283 & 0.9844 & 35.9620 & 3.7969 & 0.9503 & 32.7644 & 4.2651 & 0.9304 & 30.5036 \\
FusionNet (resize)~\cite{ref22} & 6.6084 & 0.9605 & 34.7492 & 4.6387 & 0.8038 & 29.1078 & 5.0445 & 0.7567 & 27.6059 \\
SP+M+I Net (resize)~\cite{ref02} & 6.2019 & 0.9774 & 35.7971 & 4.3615 & 0.9071 & 30.8697 & 4.7650 & 0.8800 & 29.1754 \\
DHAN~\cite{ref18} & 8.6784 & 0.9833 & 32.4535 & 6.7880 & 0.9433 & 26.2158 & 7.2224 & 0.9244 & 24.8628 \\
InstanceShadowDiffusion~\cite{ref34} & 5.0735 & \textbf{0.9865} & 37.7354 & 3.1269 & 0.9514 & 34.3006 & 3.5342 & 0.9356 & 32.1079 \\
SFR-Net (Ours) & \textbf{4.9750} & \textbf{0.9865} & \underline{37.8747} & 3.0631 & \underline{0.9550} & \underline{34.5793} & \underline{3.4627} & 0.9382 & \underline{32.3872} \\
\bottomrule
\end{tabular}%
}
\par\smallskip
\begin{minipage}{\textwidth}
\footnotesize Bold and underlined values indicate the best and second-best performances for each metric, respectively.
\end{minipage}
\endgroup
\end{table*}

\subsection{Comparison with State-of-the-Art Models}
\label{subsec:comparison}

We compared the proposed SFR-Net with state-of-the-art supervised shadow removal techniques. Performance evaluations were conducted for the following models: ShadowDiffusion~\cite{ref33}, HomoFormer~\cite{ref30}, Inpaint4Shadow~\cite{ref35}, ShadowFormer~\cite{ref10}, BMNet~\cite{ref01}, SG-ShadowNet~\cite{ref11}, FusionNet~\cite{ref22}, SP+M+I Net~\cite{ref02}, DHAN~\cite{ref18}, and InstanceShadowDiffusion~\cite{ref34}. For each model, the official results of the respective studies were used. The evaluation code used was the MATLAB code provided in~\cite{ref37}, which is widely used to evaluate the quality of shadow-removed images. In our experiments, the size of the output images of the models was set to $640 \times 480$, which is the standard size for the ISTD+ dataset.

\noindent\textbf{Quantitative evaluation.} Table~\ref{tab:tab1} lists the quantitative performance evaluation results of the various shadow removal models on the ISTD+ dataset. Because the SP+M+I Net and FusionNet models generated $256 \times 256$ images, we resized them to $640 \times 480$ images according to the ISTD+ dataset before evaluating their performances. For the shadow region, the performance of SFR-Net was similar to or better than that of existing supervised shadow removal models. Among the recent diffusion-based approaches, InstanceShadowDiffusion~\cite{ref34} demonstrates competitive performance, particularly in preserving structural details, whereas ShadowDiffusion shows strengths in color consistency. Compared to HomoFormer, an existing state-of-the-art model, SFR-Net reduced the RMSE for the shadow region from 5.0098 to 4.9750. Furthermore, SFR-Net exhibited the second-best performance regarding PSNR for the shadowed region, SSIM and PSNR for the non-shadowed region, and RMSE and PSNR for the entire region. Overall, the proposed SFR-Net outperformed the existing models in the shadow region and exhibited outstanding performance in the non-shadow region, which was comparable to its performance in the shadow region. Compared to state-of-the-art shadow removal methods, the output of SFR-Net demonstrated considerable image reconstruction performance in the non-shadow region. However, this performance was relatively low compared with that in the shadow region. This is because SFR-Net generates images while preserving the maximum information in the non-shadow region of the original shadow image $I_S$ through postprocessing. The image-inpainting technique reconstructs the region outside the mask exactly the same as the input image because there is no difference between the damaged image and the GT image in the region outside the mask. However, as listed in Table~\ref{tab:tab1}, a shadow dataset similar to the ISTD+ dataset exhibited an average RMSE of 3.1493, compared with the differences in the non-shadow region between $I_S$ and $I_{gt}$. This result shows that there are differences in the pixel values in the non-shadowed region between $I_S$ and $I_{gt}$. Therefore, the proposed SFR-Net postprocessing restores colors similar to those in the non-shadow region of $I_S$, leading to more significant differences compared with $I_{gt}$.

\begin{figure*}[!t]
    \centering
    \begin{subfigure}[b]{0.12\textwidth}
        \centering
        \includegraphics[width=\linewidth,viewport=0 0 62 243,clip]{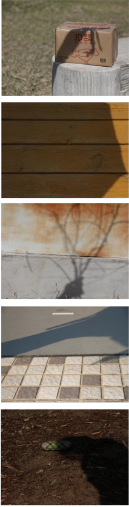}
        \caption{}\label{fig:fig5a}
    \end{subfigure}\hfill
    \begin{subfigure}[b]{0.12\textwidth}
        \centering
        \includegraphics[width=\linewidth,viewport=64 0 126 243,clip]{figures/fig5.pdf}
        \caption{}\label{fig:fig5b}
    \end{subfigure}\hfill
    \begin{subfigure}[b]{0.12\textwidth}
        \centering
        \includegraphics[width=\linewidth,viewport=129 0 191 243,clip]{figures/fig5.pdf}
        \caption{}\label{fig:fig5c}
    \end{subfigure}\hfill
    \begin{subfigure}[b]{0.12\textwidth}
        \centering
        \includegraphics[width=\linewidth,viewport=193 0 255 243,clip]{figures/fig5.pdf}
        \caption{}\label{fig:fig5d}
    \end{subfigure}\hfill
    \begin{subfigure}[b]{0.12\textwidth}
        \centering
        \includegraphics[width=\linewidth,viewport=257 0 319 243,clip]{figures/fig5.pdf}
        \caption{}\label{fig:fig5e}
    \end{subfigure}\hfill
    \begin{subfigure}[b]{0.12\textwidth}
        \centering
        \includegraphics[width=\linewidth,viewport=321 0 383 243,clip]{figures/fig5.pdf}
        \caption{}\label{fig:fig5f}
    \end{subfigure}\hfill
    \begin{subfigure}[b]{0.12\textwidth}
        \centering
        \includegraphics[width=\linewidth,viewport=386 0 448 243,clip]{figures/fig5.pdf}
        \caption{}\label{fig:fig5g}
    \end{subfigure}\hfill
    \begin{subfigure}[b]{0.12\textwidth}
        \centering
        \includegraphics[width=\linewidth,viewport=450 0 512 243,clip]{figures/fig5.pdf}
        \caption{}\label{fig:fig5h}
    \end{subfigure}

    \caption{Qualitative evaluation results of SFR-Net and state-of-the-art
    shadow removal methods for the SRD dataset. (a) Input,
    (b) ShadowDiffusion~\cite{ref33}, (c) HomoFormer~\cite{ref30},
    (d) ShadowFormer~\cite{ref10}, (e) BMNet~\cite{ref01},
    (f) Ours, (g) InstanceShadowDiffusion~\cite{ref34}, (h) GT.}
    \label{fig:fig5}
\end{figure*}
\begin{table*}[t]
\centering
\caption{Quantitative evaluation results of SFR-Net and state-of-the-art shadow removal methods on the SRD dataset.}
\label{tab:tab2}
\begingroup
\setlength{\tabcolsep}{4pt}
\renewcommand{\arraystretch}{1.15}
\small
\resizebox{\textwidth}{!}{%
\begin{tabular}{lccccccccc}
\toprule
\multicolumn{1}{c}{\multirow{2}{*}[-0.8ex]{Method}} & \multicolumn{3}{c}{Shadow Region} & \multicolumn{3}{c}{Non-Shadow Region} & \multicolumn{3}{c}{All} \\
\cmidrule(lr){2-4} \cmidrule(lr){5-7} \cmidrule(lr){8-10}
 & RMSE$\downarrow$ & SSIM$\uparrow$ & PSNR$\uparrow$ & RMSE$\downarrow$ & SSIM$\uparrow$ & PSNR$\uparrow$ & RMSE$\downarrow$ & SSIM$\uparrow$ & PSNR$\uparrow$ \\
\midrule
Shadow Image & 17.3859 & 0.8694 & 18.7869 & 2.7657 & 0.9683 & 30.2704 & 16.3193 & 0.8206 & 18.0297 \\
ShadowDiffusion~\cite{ref33} & 4.1636 & 0.9662 & 31.7264 & 2.5340 & 0.9651 & 31.2735 & 5.3763 & 0.9222 & 27.8394 \\
HomoFormer~\cite{ref30} & \textbf{3.4217} & 0.9676 & 33.1020 & 2.2344 & 0.9502 & 31.9545 & 4.5635 & 0.8988 & 28.8686 \\
Inpaint4Shadow~\cite{ref35} & 3.8828 & 0.9713 & 32.2222 & 2.1370 & 0.9739 & 32.7109 & 4.8368 & 0.9324 & 28.7194 \\
ShadowFormer~\cite{ref10} & 3.5833 & \underline{0.9739} & 33.0057 & \textbf{1.9617} & \textbf{0.9758} & 33.7010 & 4.4366 & \textbf{0.9374} & \underline{29.6462} \\
BMNet~\cite{ref01} & 3.6952 & 0.9656 & 32.3086 & 2.3587 & 0.9449 & 31.0833 & 4.8848 & 0.8889 & 27.9986 \\
SG-ShadowNet~\cite{ref11} & 3.7124 & 0.9733 & 32.5460 & \underline{1.9659} & \underline{0.9748} & \textbf{33.9631} & 4.4926 & 0.9340 & 29.6179 \\
FusionNet (resize)~\cite{ref22} & 4.2753 & 0.9569 & 31.4184 & 2.9092 & 0.9277 & 29.7585 & 5.8187 & 0.8606 & 26.9399 \\
InstanceShadowDiffusion~\cite{ref34} & 6.1959 & 0.9705 & \textbf{34.1167} & 3.6498 & 0.9611 & 32.4489 & \underline{4.4195} & 0.9259 & 29.4738 \\
DHAN~\cite{ref18} & 3.9867 & 0.9622 & 31.6425 & 2.5106 & 0.9405 & 30.6378 & 5.2274 & 0.8821 & 27.5300 \\
SFR-Net (Ours) & \underline{3.5184} & \textbf{0.9744} & \underline{33.1909} & 1.9665 & 0.9727 & \underline{33.7399} & \textbf{4.3781} & \underline{0.9341} & \textbf{29.8287} \\
\bottomrule
\end{tabular}%
}
\par\smallskip
\begin{minipage}{\textwidth}
\end{minipage}
\endgroup
\end{table*}

\begin{table}[!t]
\centering
\caption{Quantitative evaluation results of SFR-Net and state-of-the-art shadow removal methods for the non-shadow region of the original shadow image $I_S$ on the ISTD+ and SRD datasets.}
\label{tab:tab3}
\begingroup
\setlength{\tabcolsep}{4pt}
\renewcommand{\arraystretch}{1.15}
\small
\resizebox{\columnwidth}{!}{%
\begin{tabular}{lcccccc}
\toprule
\multicolumn{1}{c}{\multirow{2}{*}[-0.8ex]{Method}} & \multicolumn{3}{c}{ISTD+ Dataset} & \multicolumn{3}{c}{SRD Dataset} \\
\cmidrule(lr){2-4} \cmidrule(lr){5-7}
 & RMSE$\downarrow$ & SSIM$\uparrow$ & PSNR$\uparrow$ & RMSE$\downarrow$ & SSIM$\uparrow$ & PSNR$\uparrow$ \\
\midrule
ShadowDiffusion~\cite{ref33} & 2.495 & 0.9424 & 34.4869 & 2.7020 & \underline{0.9755} & 30.3506 \\
HomoFormer~\cite{ref30} & 1.6334 & 0.9863 & 38.7659 & 3.0385 & 0.9457 & 28.8426 \\
Inpaint4Shadow~\cite{ref35} & 3.8885 & 0.8237 & 30.2485 & 2.5566 & 0.9617 & 30.0065 \\
ShadowFormer~\cite{ref10} & \underline{1.2962} & 0.9909 & 41.0936 & \underline{2.4080} & 0.9649 & \underline{30.6063} \\
BMNet~\cite{ref01} & 1.4613 & 0.9928 & 40.0063 & 2.6261 & 0.9633 & 29.9894 \\
SG-ShadowNet~\cite{ref11} & 2.5336 & 0.9889 & 36.6816 & 2.5368 & 0.9633 & 30.5528 \\
FusionNet (resize)~\cite{ref22} & 4.4263 & 0.8033 & 29.1979 & 3.4363 & 0.9254 & 28.0174 \\
SP+M+I Net (resize)~\cite{ref02} & 3.9765 & 0.9199 & 31.1807 & -- & -- & -- \\
DHAN~\cite{ref18} & 6.1279 & 0.9824 & 27.5092 & 2.8048 & 0.95852 & 29.4690 \\
InstanceShadowDiffusion~\cite{ref34} & \textbf{1.2365} & \underline{0.9933} & \textbf{42.9716} & 4.9513 & 0.9641 & 29.4994 \\
SFR-Net (Ours) & 1.6281 & \textbf{0.9941} & \underline{41.8093} & \textbf{2.3307} & \textbf{0.9859} & \textbf{32.8520} \\
\bottomrule
\end{tabular}%
}
\par\smallskip
\begin{minipage}{\textwidth}
\end{minipage}
\endgroup
\end{table}


To evaluate the generalization capability of the proposed model, additional experiments were conducted on the SRD dataset, which posed unique challenges because of the absence of official ground-truth mask annotations. Following the common practice in the field, we employed shadow masks predicted by the DHAN model~\cite{ref18} for both training and evaluation. Despite these limitations, our model achieved competitive performance in SRD. As listed in Table~\ref{tab:tab2}, SFR-Net recorded an RMSE of 3.5184 for the shadow regions and 4.3781 for the entire image, demonstrating a performance comparable to that of the state-of-the-art methods. The slightly lower performance compared to the ISTD+ dataset can be attributed to two main factors: (1) the inherent inaccuracies in the predicted masks, which affect both the shadow removal quality and evaluation metrics, and (2) the inability to generate adjusted ground truth images for SRD, as is possible with ISTD+, which limits the accuracy of the color consistency evaluation.

To further assess the robustness of the model in preserving unaffected regions, Table~\ref{tab:tab3} presents a comparative analysis of the reconstruction performance in the non-shadow regions of the original shadow images across both the ISTD+ and SRD datasets. The results show that SFR-Net maintains high fidelity in these areas, achieving a competitive performance across different metrics. For the ISTD+ dataset, InstanceShadowDiffusion achieved the best RMSE (1.2365) and PSNR (42.9716), whereas SFR-Net demonstrated the highest SSIM (0.9941), indicating superior structural similarity preservation. On the SRD dataset, SFR-Net achieved the best performance regarding both RMSE (2.3307) and PSNR (32.8520) while maintaining the highest SSIM (0.9859). Notably, SFR-Net consistently ranked among the top performers across both datasets, demonstrating its effectiveness in preserving structural and perceptual quality against diverse baseline methods ranging from traditional CNN-based approaches to modern diffusion and transformer architectures. These findings reinforce the generalizability of our approach under diverse and less-controlled conditions.

\noindent\textbf{Qualitative evaluation.} Figure~\ref{fig:fig4} shows representative examples of shadow removal using various models, including SFR-Net, on the ISTD+ dataset, and Figure~\ref{fig:fig5} shows the corresponding results for the SRD dataset. Both figures allow a visual comparison of shadow-removed images across the models. Each figure includes images of varying complexity: simple scenes (first and second rows) and more complex scenes (third and fifth rows). Among the shadow-removed images, traces of shadow removal were generally more noticeable in simple than in complex visual images. When attempting to remove shadows in simple images, similar to the first and second rows of Figures~\ref{fig:fig4} and~\ref{fig:fig5}, the output of SFR-Net closely resembles the ground truth compared to the results of the other models and does not leave any traces. Even in more complex images, the output of SFR-Net leaves fewer traces of shadow removal than other shadow removal models. From a color comparison within the shadowed region, we observed that the colors inside the shadowed region were restored in a manner similar to that in the GT image.

\begin{figure*}[!t]
    \centering
    \includegraphics[width=\linewidth]{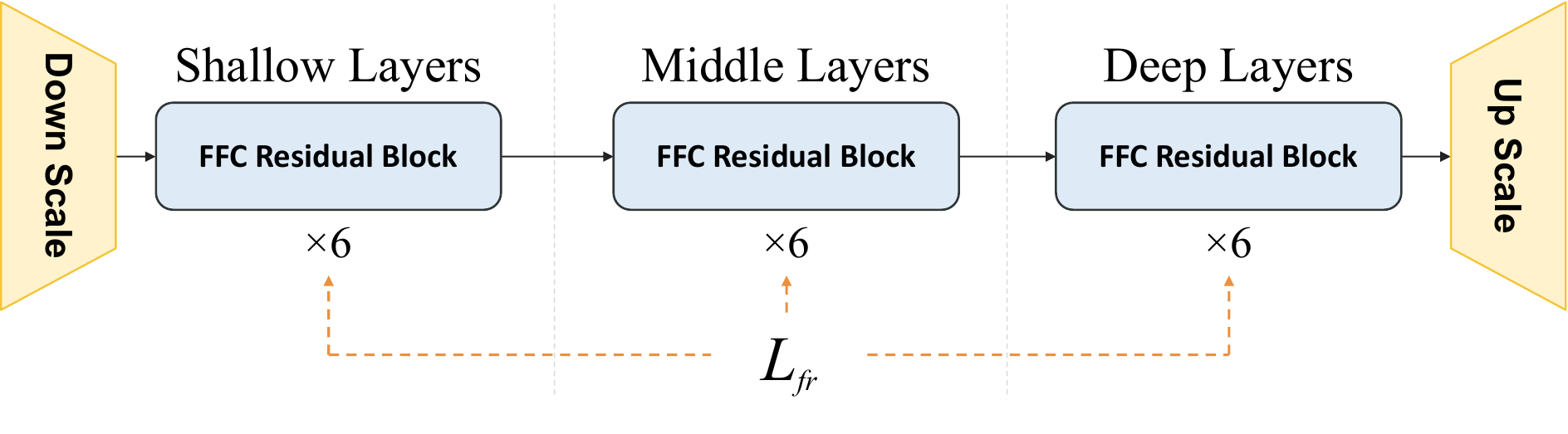}
    \caption{Division of layers into three stages for analyzing the effectiveness of knowledge distillation}
    \label{fig:fig6}
\end{figure*}
\begin{figure*}[!t]
    \centering
    \begingroup
    \setlength{\tabcolsep}{1pt}
    \renewcommand{\arraystretch}{0}

    \newcommand{\figtwoimg}[1]{%
        \includegraphics[
            width=\dimexpr(\linewidth-14pt)/4\relax
        ]{figures/fig7/fig7_#1.pdf}%
    }

    \begin{tabular}{@{}*{4}{c}@{}}
        \figtwoimg{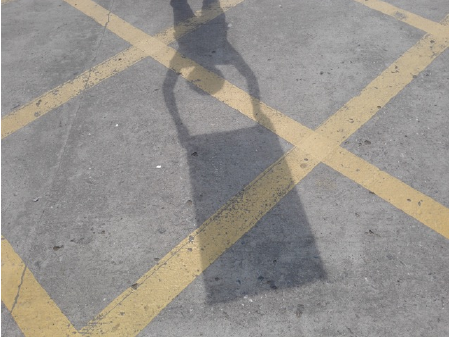} & \figtwoimg{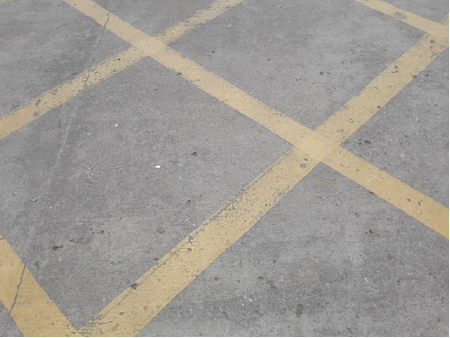} &
        \figtwoimg{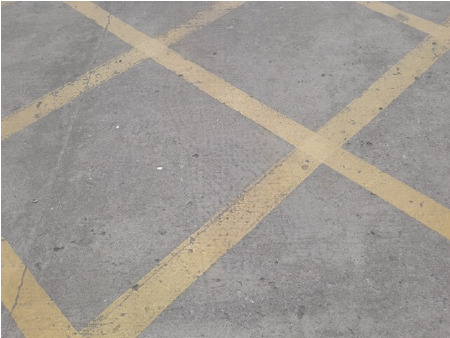} & \figtwoimg{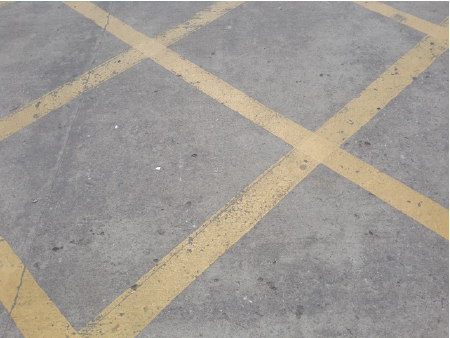} \\[2pt]

        \figtwoimg{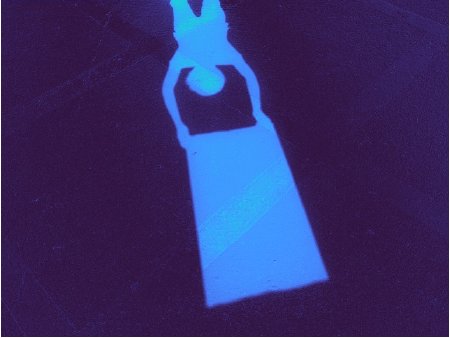} & \figtwoimg{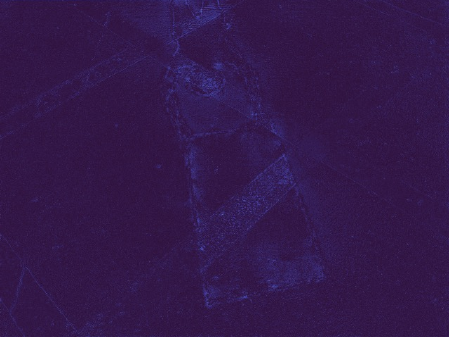} &
        \figtwoimg{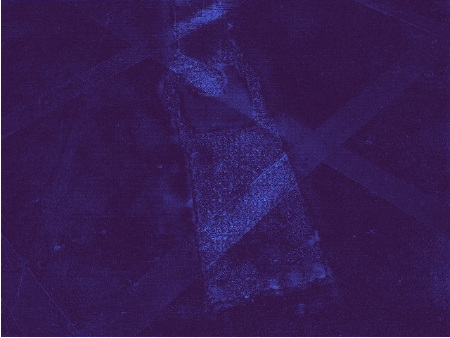} & \figtwoimg{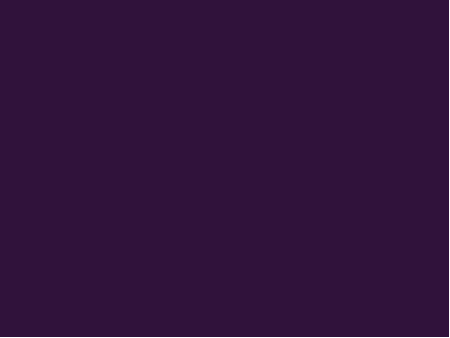} \\[2pt]

        \rule{0pt}{2.5ex}(a) & (b) & (c) & (d) \\
    \end{tabular}
    \endgroup

    \caption{Matte images when $L_{fr}$ is applied to SFR-Net and when $L_{fr}$ is not applied to SFR-Net. (a) $I_s$, (b) With $L_{fr}$, (c) Without $L_{fr}$, (d) $I_{gt}$}
    \label{fig:fig7}
\end{figure*}
\begin{table}[!t]
\centering
\caption{Evaluation of shadow removal performance when feature refinement loss is applied to each stage of SFR-Net.}
\label{tab:tab4}
\begingroup
\setlength{\tabcolsep}{4pt}
\renewcommand{\arraystretch}{1.15}
\small
\resizebox{\columnwidth}{!}{%
\begin{tabular}{lccccccccc}
\toprule
\multicolumn{1}{c}{\multirow{2}{*}[-0.8ex]{Method}} & \multicolumn{3}{c}{Shadow Region} & \multicolumn{3}{c}{Non-Shadow Region} & \multicolumn{3}{c}{All} \\
\cmidrule(lr){2-4} \cmidrule(lr){5-7} \cmidrule(lr){8-10}
 & RMSE$\downarrow$ & SSIM$\uparrow$ & PSNR$\uparrow$ & RMSE$\downarrow$ & SSIM$\uparrow$ & PSNR$\uparrow$ & RMSE$\downarrow$ & SSIM$\uparrow$ & PSNR$\uparrow$ \\
\midrule
\shortstack[l]{Config.\ \#1\\(Shallow Only)} & 6.0479 & 0.9852 & 35.9776 & 3.6449 & 0.9515 & 32.5604 & 4.1722 & 0.9332 & 30.2424 \\
\shortstack[l]{Config.\ \#2\\(Middle Only)} & 5.5084 & \underline{0.9858} & 36.8616 & 3.4796 & \underline{0.9542} & 33.7787 & 3.9229 & \underline{0.9369} & 31.4653 \\
\shortstack[l]{Config.\ \#3\\(Deep Only)} & \textbf{5.2107} & \textbf{0.9862} & \textbf{37.4321} & \textbf{3.2502} & \underline{0.9542} & \textbf{34.0711} & \textbf{3.6675} & \textbf{0.9372} & \textbf{31.8471} \\
\shortstack[l]{Config.\ \#4\\(Shallow \& Middle)} & 5.6130 & 0.9849 & 36.5651 & \underline{3.3170} & \textbf{0.9545} & 33.6416 & 3.8253 & 0.9360 & 31.3321 \\
\shortstack[l]{Config.\ \#5\\(Middle \& Deep)} & 5.4761 & 0.9857 & \underline{37.1288} & 3.3211 & 0.9518 & \underline{34.0083} & \underline{3.7688} & 0.9341 & \underline{31.7260} \\
\shortstack[l]{Config.\ \#6\\(All Stages)} & \underline{5.4430} & 0.9854 & 36.8779 & 3.3861 & 0.9540 & 33.6865 & 3.8286 & 0.9361 & 31.3961 \\
Without $L_{fr}$ & 6.0975 & 0.9849 & 35.4867 & 3.7747 & 0.9533 & 31.7673 & 4.2855 & 0.9347 & 29.6694 \\
\bottomrule
\end{tabular}%
}
\par\smallskip
\begin{minipage}{\textwidth}
\end{minipage}
\endgroup
\end{table}

\begin{table}[t]
\centering
\caption{Evaluation of shadow removal performance of SFR-Net when $L_{LaMa}$, $L_{mix}$, and $L_{fr}$ are applied.}
\label{tab:tab5}
\begingroup
\setlength{\tabcolsep}{4pt}
\renewcommand{\arraystretch}{1.15}
\small
\resizebox{\linewidth}{!}{%
\begin{tabular}{cccccccccccc}
\toprule
\multicolumn{3}{c}{Method} & \multicolumn{3}{c}{Shadow Region} & \multicolumn{3}{c}{Non-Shadow Region} & \multicolumn{3}{c}{All} \\
\cmidrule(lr){4-6} \cmidrule(lr){7-9} \cmidrule(lr){10-12} \cmidrule(lr){1-3}
$L_{LaMa}$ & $L_{mix}$ & $L_{fr}$ & RMSE$\downarrow$ & SSIM$\uparrow$ & PSNR$\uparrow$ & RMSE$\downarrow$ & SSIM$\uparrow$ & PSNR$\uparrow$ & RMSE$\downarrow$ & SSIM$\uparrow$ & PSNR$\uparrow$ \\
\midrule
$\checkmark$ &  &  & 7.4366 & 0.9774 & 34.0190 & 5.1019 & 0.9249 & 29.7530 & 5.6161 & 0.8976 & 27.8767 \\
$\checkmark$ & $\checkmark$ &  & \underline{6.0975} & \underline{0.9849} & 35.4867 & \underline{3.7747} & \underline{0.9533} & \underline{31.7673} & \underline{4.2855} & \underline{0.9347} & 29.6694 \\
$\checkmark$ &  & $\checkmark$ & 6.4203 & 0.9827 & \underline{35.6343} & 4.1685 & 0.9472 & 31.5645 & 4.6311 & 0.9263 & \underline{29.7403} \\
$\checkmark$ & $\checkmark$ & $\checkmark$ & \textbf{5.2107} & \textbf{0.9862} & \textbf{37.4321} & \textbf{3.2502} & \textbf{0.9542} & \textbf{34.0711} & \textbf{3.6675} & \textbf{0.9372} & \textbf{31.8471} \\
\bottomrule
\end{tabular}%
}
\par\smallskip
\begin{minipage}{\textwidth}
\end{minipage}
\endgroup
\end{table}


\subsection{Ablation Study}
\label{subsec:ablation}

To assess the effectiveness of knowledge distillation, we divided the SFR-Net layers into three stages: shallow, middle, and deep. We further analyzed the performance changes based on whether the feature refinement loss was applied at each stage. Figure~\ref{fig:fig6} shows the divisions of each stage. The experiments enabled a comparison of the performance of applying $L_{fr}$ loss to one of the three stages (Config.~\#1, Config.~\#2, and Config.~\#3), applying $L_{fr}$ loss to two stages (Config.~\#4 and Config.~\#5), applying $L_{fr}$ loss to all three stages (Config.~\#6), and not applying $L_{fr}$ loss to any stage (Config.~\#7). In all cases, the performance was better when feature refinement loss was applied than when it was not. Furthermore, according to the experimental results, the performance improved when feature refinement loss was applied to the deep layer, regardless of the number of stages in which the feature refinement loss was applied. Our hypothesis is that as SFR-Net removes shadows, it learns through knowledge distillation and gradually generates features that are similar to those of the GT image. Therefore, we assumed that better results would be obtained when the feature refinement loss was applied to the deep layer than when it was applied to the shallow layer. Specifically, shadow images $I_S$ and binary masks $M_S$ are used as inputs for shadow removal. As the layers deepened, the boundary between the two images gradually deteriorated. In the initial layers, the features of $I_S$ and $M_S$ do not converge well. Therefore, applying feature refinement loss to these layers may not significantly improve performance. However, in the deep layer, the features of the mask and shadow images converge more effectively. Hence, performance can be significantly improved through feature refinement using GT images. The experimental results, as listed in Table~\ref{tab:tab4}, support our hypothesis that shadow removal performance improves when knowledge distillation is applied closer to the latter stages.

We further compared the performance of Config.~\#3, in which we observed the best shadow removal performance, depending on the application of individual losses (see Table~\ref{tab:tab5}). Specifically, $L_{LaMa}$ is the loss used in LaMa~\cite{ref36}, the SFR-Net base model, and includes the $L_{Adv}$, $L_{HRFPL}$, $L_{fm}$, and $L_1$ loss, which are reconstruction losses. $L_{mix}$ is the reconstruction loss that combines the $L_1$ loss and MS-SSIM, whereas $L_{fr}$ represents the feature refinement loss.

\begin{table*}[t]
\centering
\caption{Quantitative comparison of different post-processing techniques on the ISTD+ dataset.}
\label{tab:tab6}
\begingroup
\setlength{\tabcolsep}{4pt}
\renewcommand{\arraystretch}{1.15}
\small
\resizebox{\textwidth}{!}{%
\begin{tabular}{lcccccccccc}
\toprule
Method & \multicolumn{3}{c}{Shadow Region} & \multicolumn{3}{c}{Non-Shadow Region} & \multicolumn{3}{c}{All} & \shortstack{Processing Time\\(second)} \\
\cmidrule(lr){2-4} \cmidrule(lr){5-7} \cmidrule(lr){8-10}
 & RMSE$\downarrow$ & SSIM$\uparrow$ & PSNR$\uparrow$ & RMSE$\downarrow$ & SSIM$\uparrow$ & PSNR$\uparrow$ & RMSE$\downarrow$ & SSIM$\uparrow$ & PSNR$\uparrow$ &  \\
\midrule
SFR-Net (default) & 5.2107 & 0.9862 & 37.4321 & 3.2502 & 0.9542 & 34.0711 & 3.6675 & 0.9372 & 31.8471 & -- \\
\midrule
\multicolumn{11}{l}{\textbf{BGR Space}} \\
+ linear regression & \textbf{4.9750} & \textbf{0.9865} & \textbf{37.8747} & \underline{3.0631} & 0.9550 & \underline{34.5793} & \underline{3.4627} & \underline{0.9382} & \underline{32.3872} & 0.068 \\
+ polynomial regression & 5.2444 & 0.9860 & 37.6492 & 3.2386 & \underline{0.9553} & 34.3712 & 3.6599 & 0.9380 & 32.1919 & 0.187 \\
+ histogram matching & 10.8398 & 0.9698 & 32.9278 & 3.2208 & 0.9541 & 34.2324 & 5.0366 & 0.9185 & 29.9032 & \underline{0.050} \\
+ color transfer~\cite{ref46} & 9.2768 & 0.976 & 33.6943 & 3.2208 & 0.9541 & 34.2324 & 4.5661 & 0.9260 & 30.3646 & 0.056 \\
+ gradient boosting & 5.0741 & 0.9862 & \underline{37.7767} & \textbf{2.9127} & \textbf{0.9563} & \textbf{35.0749} & \textbf{3.3700} & \textbf{0.9390} & \textbf{32.6578} & 2.513 \\
+ gamma correction & 8.4250 & 0.9806 & 34.8413 & 3.2208 & 0.9541 & 34.2324 & 4.3524 & 0.9300 & 30.9087 & \textbf{0.043} \\
\midrule
\multicolumn{11}{l}{\textbf{LAB Space}} \\
+ linear regression & \textbf{5.0013} & \underline{0.9857} & \textbf{37.7963} & \underline{3.1621} & 0.9505 & \underline{34.3575} & \underline{3.5482} & \underline{0.9326} & \underline{32.2254} & \underline{0.054} \\
+ polynomial regression & 5.1281 & 0.9853 & 37.6323 & 3.3148 & 0.9501 & 34.0300 & 3.6971 & 0.9319 & 31.9582 & 0.158 \\
+ histogram matching & 11.5396 & 0.9672 & 32.3056 & 3.2834 & \underline{0.9523} & 34.1094 & 5.2168 & 0.9139 & 29.5462 & 0.059 \\
+ color transfer~\cite{ref46} & 9.4752 & 0.9755 & 33.5719 & 3.2834 & \underline{0.9523} & 34.1094 & 4.6691 & 0.9227 & 30.2466 & 0.059 \\
+ gradient boosting & 5.0753 & \textbf{0.9862} & \underline{37.7766} & \textbf{2.9127} & \textbf{0.9562} & \underline{35.0753} & \textbf{3.3712} & \textbf{0.9390} & \textbf{32.6574} & 1.937 \\
+ gamma correction & 8.5674 & 0.9792 & 34.7292 & 3.2834 & \underline{0.9523} & 34.1094 & 4.4338 & 0.9267 & 30.7879 & \textbf{0.048} \\
\bottomrule
\end{tabular}%
}
\par\smallskip
\begin{minipage}{\textwidth}
\end{minipage}
\endgroup
\end{table*}

\begin{table*}[t]
\centering
\caption{Quantitative comparison of different post-processing techniques on the SRD dataset.}
\label{tab:tab7}
\begingroup
\setlength{\tabcolsep}{4pt}
\renewcommand{\arraystretch}{1.15}
\small
\resizebox{\textwidth}{!}{%
\begin{tabular}{lcccccccccc}
\toprule
Method & \multicolumn{3}{c}{Shadow Region} & \multicolumn{3}{c}{Non-Shadow Region} & \multicolumn{3}{c}{All} & \shortstack{Processing Time\\(second)} \\
\cmidrule(lr){2-4} \cmidrule(lr){5-7} \cmidrule(lr){8-10}
 & RMSE$\downarrow$ & SSIM$\uparrow$ & PSNR$\uparrow$ & RMSE$\downarrow$ & SSIM$\uparrow$ & PSNR$\uparrow$ & RMSE$\downarrow$ & SSIM$\uparrow$ & PSNR$\uparrow$ &  \\
\midrule
SFR-Net (default) & 3.7840 & 0.9719 & 32.9293 & 2.5129 & 0.9690 & 33.5797 & 5.2807 & 0.9266 & 29.4810 & -- \\
\midrule
\multicolumn{11}{l}{\textbf{BGR Space}} \\
+ linear regression & \underline{3.5184} & \textbf{0.9744} & \underline{33.1909} & \underline{1.9665} & 0.9727 & \underline{33.7399} & \underline{4.3781} & \underline{0.9341} & \underline{29.8287} & 0.046 \\
+ polynomial regression & 3.8751 & 0.9719 & 32.8250 & 2.4020 & 0.9703 & 33.6360 & 5.1542 & 0.9277 & 29.4965 & 0.146 \\
+ histogram matching & 6.8171 & 0.9445 & 28.2236 & 1.9761 & \underline{0.9743} & 33.7197 & 6.9081 & 0.9037 & 26.4355 & \underline{0.030} \\
+ color transfer~\cite{ref46} & 6.0244 & 0.9515 & 29.0468 & 1.9754 & \underline{0.9743} & 33.7206 & 6.2467 & 0.9106 & 27.0794 & 0.036 \\
+ gradient boosting & \textbf{3.4189} & \underline{0.9731} & \textbf{33.4204} & \textbf{1.7751} & \textbf{0.9754} & \textbf{34.8569} & \textbf{4.1269} & \textbf{0.9346} & \textbf{30.4812} & 2.517 \\
+ gamma correction & 5.4126 & 0.9579 & 29.9249 & 1.9748 & \underline{0.9743} & 33.7221 & 5.7732 & 0.9175 & 27.6969 & \textbf{0.022} \\
\midrule
\multicolumn{11}{l}{\textbf{LAB Space}} \\
+ linear regression & \underline{3.5340} & \underline{0.9720} & \underline{33.0943} & 2.1449 & 0.9710 & \underline{33.7517} & \underline{4.6310} & \underline{0.9288} & \underline{29.7129} & 0.052 \\
+ polynomial regression & 3.6287 & 0.9716 & 32.8703 & 2.1404 & 0.9707 & 33.7016 & 4.6761 & 0.9283 & 29.5851 & 0.158 \\
+ histogram matching & 6.8738 & 0.9438 & 28.0923 & 1.9940 & \textbf{0.9738} & 33.6939 & 6.9104 & 0.9024 & 26.3554 & \underline{0.034} \\
+ color transfer~\cite{ref46} & 6.0680 & 0.9509 & 28.9820 & 1.9938 & \textbf{0.9738} & 33.6943 & 6.2952 & 0.9094 & 27.0294 & 0.040 \\
+ gradient boosting & \textbf{3.4049} & \textbf{0.9725} & \textbf{33.3698} & \textbf{1.8277} & \underline{0.9732} & \textbf{34.6197} & \textbf{4.1727} & \textbf{0.9314} & \textbf{30.3473} & 1.979 \\
+ gamma correction & 5.4568 & 0.9556 & 29.8500 & \underline{1.9932} & \textbf{0.9738} & 33.6958 & 5.8259 & 0.9147 & 27.6307 & \textbf{0.028} \\
\bottomrule
\end{tabular}%
}
\par\smallskip
\begin{minipage}{\textwidth}
\end{minipage}
\endgroup
\end{table*}


We compared performances of using four different combinations of losses: when using only $L_{LaMa}$, when using $L_{LaMa}+L_{mix}$, when using $L_{LaMa}+L_{fr}$, and when using all three losses, $L_{LaMa}+L_{mix}+L_{fr}$. $L_{LaMa}+L_{mix}$ compared the luminance and contrast during reconstruction. Hence, it significantly reduces the color differences induced by variations in illumination in the non-shadow regions of $I_S$ and $I_{gt}$ in the ISTD+ dataset. In contrast, $L_{LaMa}+L_{fr}$ exhibited relatively poor performance in the non-shadowed region compared to $L_{LaMa}+L_{mix}$. However, it demonstrated superior PSNR performance in the entire shadow region, as well as in other regions. Finally, when all three losses were applied (which combined the two advantages) to Config.~\#3, SFR-Net achieved the best performance in all regions and categories.

Although deep learning models often function as black boxes, we attempted to gain insight into the internal decision process of the proposed SFR-Net by analyzing intermediate feature representations and their role in enhancing shadow removal performance. Figure~\ref{fig:fig3} shows the feature maps extracted from the 20th FFC block, with and without the application of the proposed feature refinement loss ($L_{fr}$). The maps with $L_{fr}$ exhibited more spatially coherent and texturally rich patterns, suggesting that the model learned to better discriminate between the shadowed and non-shadowed regions. This indicates that $L_{fr}$ enhances semantic abstraction in the deeper layers, thereby facilitating a more context-aware restoration. Table~\ref{tab:tab4} quantitatively supports this interpretation by showing significant performance gains when $L_{fr}$ is applied to deeper layers. Furthermore, Figure~\ref{fig:fig7} provides matte visualizations of the residuals between the ground truth and generated outputs. These visualizations highlight that when $L_{fr}$ is employed, the residual regions corresponding to the shadows become noticeably smaller and more consistent with the ground truth structure, further illustrating the model's capacity to produce accurate and explainable predictions. Taken together, these visual and quantitative analyses offer an interpretable explanation for why the proposed model effectively performs shadow removal by refining the internal features to better distinguish and reconstruct shadowed regions.

\begin{figure*}[!t]
    \centering
    \begin{subfigure}[b]{0.35\textwidth}
        \centering
        \includegraphics[width=\linewidth,
            viewport=0 0 402 239,clip]{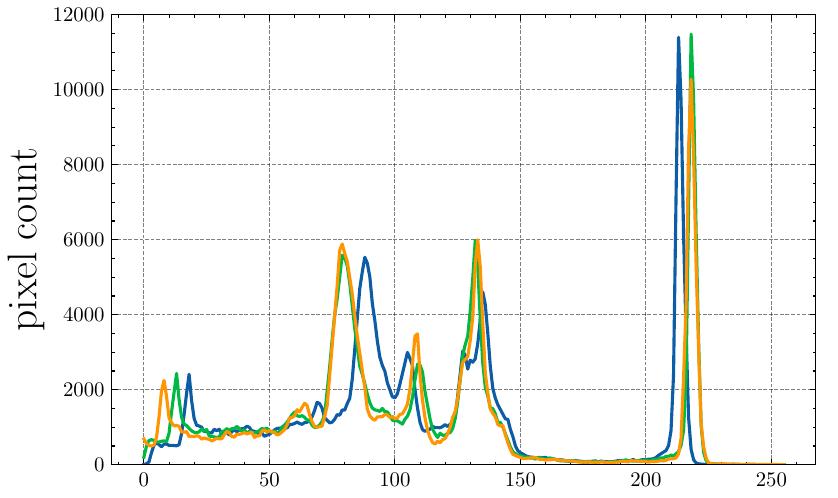}
        \caption{}\label{fig:fig8a}
    \end{subfigure}\hfill
    \begin{subfigure}[b]{0.305\textwidth}
        \centering
        \includegraphics[width=\linewidth,
            viewport=402 0 755 239,clip]{figures/fig8/fig8-abc.pdf}
        \caption{}\label{fig:fig8b}
    \end{subfigure}\hfill
    \begin{subfigure}[b]{0.305\textwidth}
        \centering
        \includegraphics[width=\linewidth,
            viewport=755 0 1105 239,clip]{figures/fig8/fig8-abc.pdf}
        \caption{}\label{fig:fig8c}
    \end{subfigure}

    \par\medskip

    \begin{subfigure}[b]{0.24\textwidth}
        \centering
        \includegraphics[width=\linewidth]{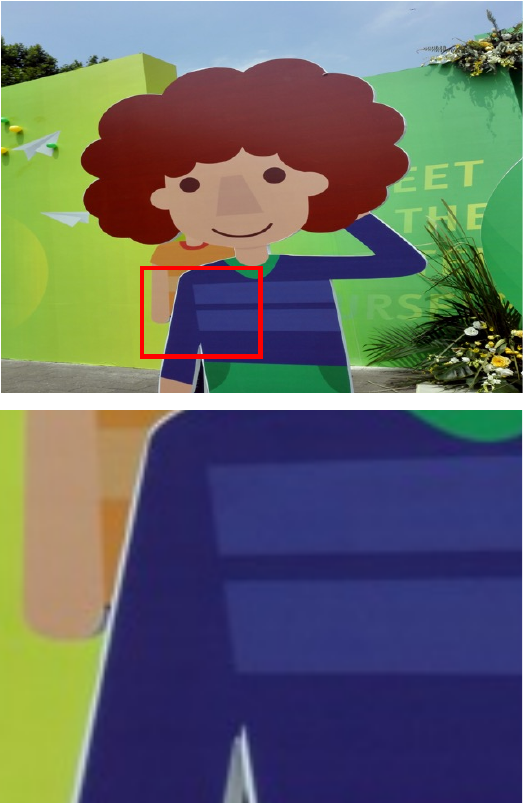}
        \caption{}\label{fig:fig8d}
    \end{subfigure}\hfill
    \begin{subfigure}[b]{0.24\textwidth}
        \centering
        \includegraphics[width=\linewidth]{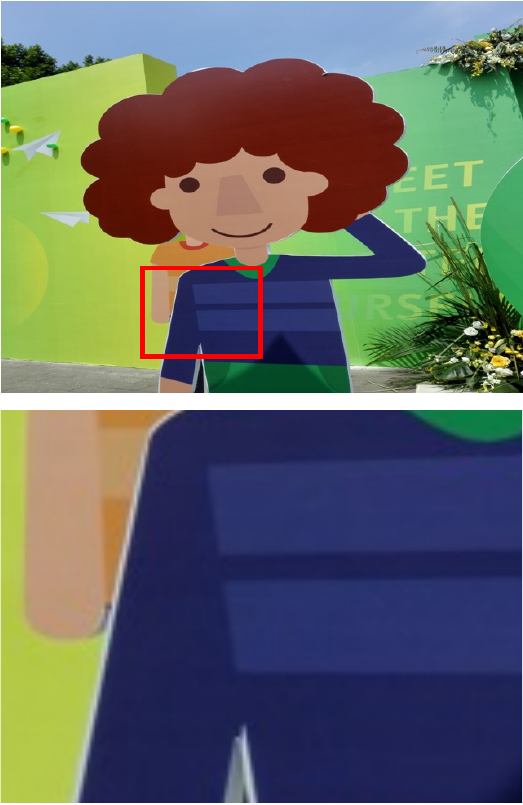}
        \caption{}\label{fig:fig8e}
    \end{subfigure}\hfill
    \begin{subfigure}[b]{0.24\textwidth}
        \centering
        \includegraphics[width=\linewidth]{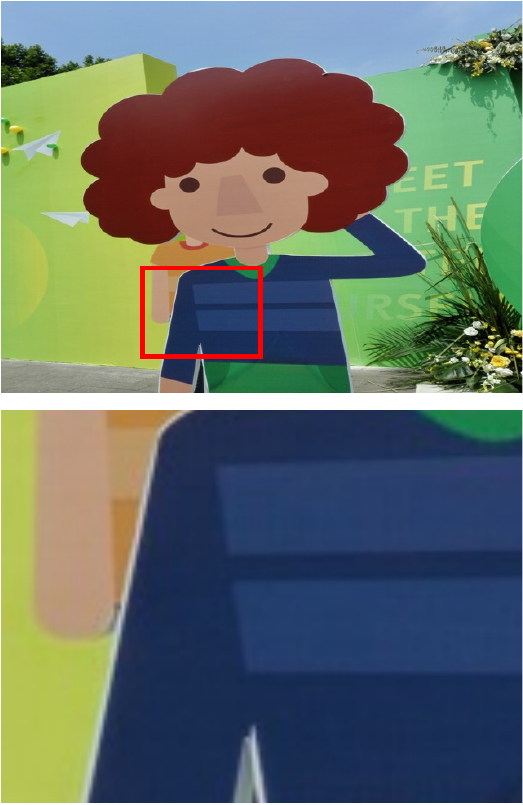}
        \caption{}\label{fig:fig8f}
    \end{subfigure}\hfill
    \begin{subfigure}[b]{0.24\textwidth}
        \centering
        \includegraphics[width=\linewidth]{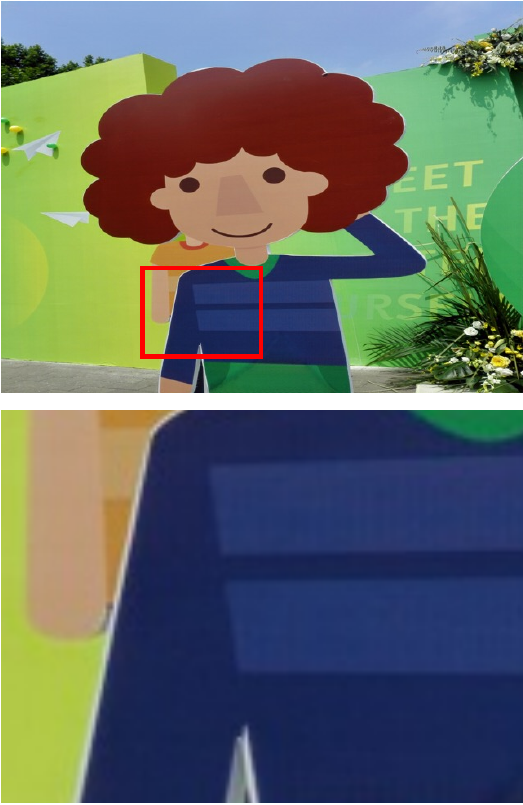}
        \caption{}\label{fig:fig8g}
    \end{subfigure}

    \caption{Comparison of color histograms when color restoration post-processing is applied to SFR-Net. (a): Blue Space Pixel Histogram, (b): Green Space Pixel Histogram, (c): Red Space Pixel Histogram, (d): $I_{gt}$, (e): $I_s$, (f): $I_{sf}$, (g): $I_{sf-pp}$}
    \label{fig:fig8}
\end{figure*}
\begin{figure*}[!t]
    \centering
    \begin{subfigure}[b]{0.14\textwidth}
        \centering
        \includegraphics[width=\linewidth,viewport=0 0 70 218,clip]{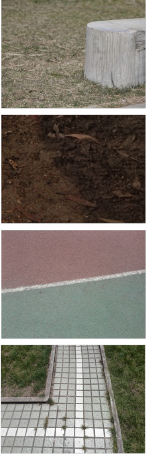}
        \caption{}\label{fig:fig9a}
    \end{subfigure}\hfill
    \begin{subfigure}[b]{0.14\textwidth}
        \centering
        \includegraphics[width=\linewidth,viewport=73 0 143 218,clip]{figures/fig9.pdf}
        \caption{}\label{fig:fig9b}
    \end{subfigure}\hfill
    \begin{subfigure}[b]{0.14\textwidth}
        \centering
        \includegraphics[width=\linewidth,viewport=145 0 215 218,clip]{figures/fig9.pdf}
        \caption{}\label{fig:fig9c}
    \end{subfigure}\hfill
    \begin{subfigure}[b]{0.14\textwidth}
        \centering
        \includegraphics[width=\linewidth,viewport=218 0 288 218,clip]{figures/fig9.pdf}
        \caption{}\label{fig:fig9d}
    \end{subfigure}\hfill
    \begin{subfigure}[b]{0.14\textwidth}
        \centering
        \includegraphics[width=\linewidth,viewport=290 0 360 218,clip]{figures/fig9.pdf}
        \caption{}\label{fig:fig9e}
    \end{subfigure}\hfill
    \begin{subfigure}[b]{0.14\textwidth}
        \centering
        \includegraphics[width=\linewidth,viewport=363 0 433 218,clip]{figures/fig9.pdf}
        \caption{}\label{fig:fig9f}
    \end{subfigure}\hfill
    \begin{subfigure}[b]{0.14\textwidth}
        \centering
        \includegraphics[width=\linewidth,viewport=435 0 505 218,clip]{figures/fig9.pdf}
        \caption{}\label{fig:fig9g}
    \end{subfigure}

    \caption{Qualitative evaluation results of post-processing algorithm for color restoration. (a): $I_{sf}$, (b): linear regression, (c): polynomial regression, (d): Reinhard et al.\cite{ref46}, (e): gradient boosting, (f): gamma correction, (g): $I_s$}
    \label{fig:fig9}
\end{figure*}

Here, $I_{sf}$, which removes shadows through the guided network $G_S$, learns the color differences in the non-shadow region of $I_S$ and applies a post-processing algorithm that improves the colors of the entire image. The greater the color difference in the non-shadowed region, the greater is the enhancement effect. When the postprocessing algorithm for color restoration was applied to the model with the best performance within SFR-Net, it improved the RMSE, SSIM, and PSNR performance in all regions. Tables~\ref{tab:tab6} and~\ref{tab:tab7} list the shadow removal performances on the ISTD+ and SRD datasets, respectively, across 12 post-processing configurations combining six algorithms (linear regression, polynomial regression, histogram matching, color transfer~\cite{ref46}, gradient boosting, and gamma correction) with two color spaces (LAB and BGR), including a baseline without post-processing. Our analysis reveals several consistent trends across both datasets. First, while all post-processing methods improve the shadow removal quality, applying them in the BGR space generally yields better results than those in the LAB space, likely owing to the improved visual coherence in the RGB channels. On the ISTD+ dataset (Table~\ref{tab:tab6}), our default method, linear regression in the BGR space, reduced the overall RMSE from 3.6675 to 3.4627 and the shadow region RMSE from 5.2107 to 4.9750, while increasing the PSNR from 37.4321 to 37.8747 and SSIM from 0.9862 to 0.9865. Similar improvements were observed for the SRD dataset (Table~\ref{tab:tab7}). Second, although gradient boosting in the BGR space achieved the highest performance metrics, it incurred a substantial computational cost of approximately 2.517~s per image, compared with 0.046~s for linear regression, representing a 54-fold increase in processing time. Therefore, for practical applications, linear regression in the BGR space provides an excellent tradeoff between performance and efficiency.

In addition, we analyzed the effectiveness of our proposed method by comparing the image $I_{sf\text{-}pp}$, obtained through the proposed linear regression-based color restoration post-processing, with the original output $I_{sf}$ before post-processing. To check for visual color differences, we plotted RGB histograms. These histograms indicate that the distribution aligned more closely with the distribution of the GT image when post-processing was used than when it was not used. Figure~\ref{fig:fig8} shows a comparison of the RGB histograms for the GT and shadow-removed images with and without linear regression-based postprocessing. Figure~\ref{fig:fig9} shows qualitative visual comparisons of these different post-processing approaches on challenging shadow images from both datasets. Linear regression produced the most natural results, with seamless integration between the previously shadowed and non-shadowed regions. In contrast, methods such as histogram matching and color transfer occasionally introduce color artifacts or oversaturated regions, despite some improvements in quantitative metrics.

\section{Conclusion}
\label{sec:conclusion}

In this study, we propose a new shadow-removal model called SFR-Net. The SFR-Net learns based on the knowledge distillation of feature maps during the shadow removal process. Specifically, the GT image was input into the pretrained image inpainting network, and the features were extracted accordingly. These features were then compared with those of the shadow removal model. Regarding the loss function, we enhanced the shadow removal performance using MS-SSIM, $L_1$ loss, and feature refinement loss. We also incorporated a method for color restoration through linear regression-based postprocessing. The experimental results indicated that SFR-Net can effectively reconstruct images for shadow removal.

However, the proposed method has several limitations. First, because the model was trained on the ISTD+ and SRD datasets, it may face challenges in generalizing to unseen environments, particularly those with significantly different lighting conditions, object types, noisy backgrounds, or complex overlapping shadows. Second, although knowledge distillation is inherently suitable for label-efficient learning, our model still depends on paired data because of the supervised nature of shadow-removal tasks. Third, although effective, the postprocessing step remains separate from the core model and cannot be optimized jointly. To address these limitations, future work will explore shadow removal models that incorporate self-supervised or weakly supervised learning schemes to reduce the dependence on paired datasets. We also aimed to develop an end-to-end architecture in which postprocessing for color correction was integrated into the network and optimized jointly, thereby improving both robustness and visual coherence. In addition, we plan to evaluate the model using more diverse datasets and challenging conditions, including varying lighting, sensor noise, and structural inconsistencies to further validate its generalization capabilities in real-world scenarios. Finally, although preliminary interpretability is provided through feature and residual visualizations, future studies may explore model attribution techniques to improve the transparency and explainability of the decision-making process.

\section{Acknowledgement}
This research was supported by the Basic Science Research Program of the National Research Foundation of Korea (NRF) funded by the Ministry of Education (RS-2020-NR049579).

\clearpage

\bibliography{aaai2027}

\end{document}